\documentclass[10pt,twocolumn,letterpaper]{article}

\usepackage[pagenumbers]{cvpr} 

\newcommand{\modelName}{A3Syn\xspace}

\usepackage{color}
\definecolor{crimson}{rgb}{0.86, 0.08, 0.24}
\definecolor{gray}{rgb}{0.5,0.5,0.5}
\definecolor{green}{rgb}{0, 0.4, 0}
\definecolor{orange}{rgb}{1, 0.5, 0}
\definecolor{mahogany}{rgb}{0.75, 0.25, 0.0}
\definecolor{purple}{rgb}{0.6, 0, 0.6}
\definecolor{darkgreen}{rgb}{0, 0.4, 0}
\definecolor{frenchblue}{rgb}{0.0, 0.45, 0.73}
\definecolor{red}{rgb}{1,0,0}
\definecolor{yellow}{rgb}{1,1,0}
\definecolor{magenta}{rgb}{1,0,1}
\definecolor{pink}{rgb}{1,0.412,0.706}
\definecolor{newgreen}{rgb}{0, 0.6, 0.2}
\definecolor{lightred}{HTML}{fff3f3}

\makeatletter
\DeclareRobustCommand\onedot{\futurelet\@let@token\@onedot}
\def\@onedot{\ifx\@let@token.\else.\null\fi\xspace}

\def\eg{\emph{e.g}\onedot} 
\def\ie{\emph{i.e}\onedot}

\makeatother

\def\eg{e.g.,~}               
\def\ie{i.e.,~}               

\newlength\paramargin
\newlength\figmargin
\newlength\subfigmargin
\newlength\presecmargin
\newlength\secmargin
\newlength\subsecmargin
\newlength\subsubsecmargin
\newlength\tabmargin
\newlength\eqmargin
\newlength\paraskip

\long\def\ignorethis#1{}

\newcommand{\ff}{\mathbf{f}}

\newcommand{\calL}{\mathcal{L}}

\newcommand{\calS}{\mathcal{S}}
\newcommand{\calC}{\mathcal{C}}

\newcommand{\issue}[1]{\vspace{0.1em}\noindent {\textbf{#1 \hspace{0.2em}}}}

\newcommand{\comment}[1]{}
\usepackage{enumitem} 
\usepackage{tikz} 
\usetikzlibrary{positioning,shapes}

\usepackage{graphicx}
\usepackage{bbm}
\usepackage{adjustbox}
\usepackage{booktabs}
\usepackage{listings}
\definecolor{cvprblue}{rgb}{0.21,0.49,0.74}
\usepackage[pagebackref,breaklinks,colorlinks,allcolors=cvprblue]{hyperref}

\def\paperID{3121} 
\def\confName{CVPR}
\def\confYear{2025}

\title{Towards Affordance-Aware Articulation Synthesis for Rigged Objects\vspace{-0.5em}}

\author{
Yu-Chu Yu$^{1}$
\,\,\,
Chieh Hubert Lin$^{2}$
\,\,\,
Hsin-Ying Lee$^{3}$
\,\,\,
Chaoyang Wang$^{3}$
\\
Yu-Chiang Frank Wang$^{1}$
\,\,\,
Ming-Hsuan Yang$^{2,4}$
\\ [0.5em]
$^{1}$ National Taiwan University
\,\,\,
$^{2}$ UC Merced
\,\,\,
$^{3}$ Snap Research
\,\,\,
$^{4}$ Yonsei University
\\ [.3em]
\url{https://chuyu.org/research/a3syn}
}

\begin{document}
\twocolumn[{%
\maketitle
\vspace{-2.0em}
\renewcommand\twocolumn[1][]{#1}%
    \includegraphics[width=\linewidth]{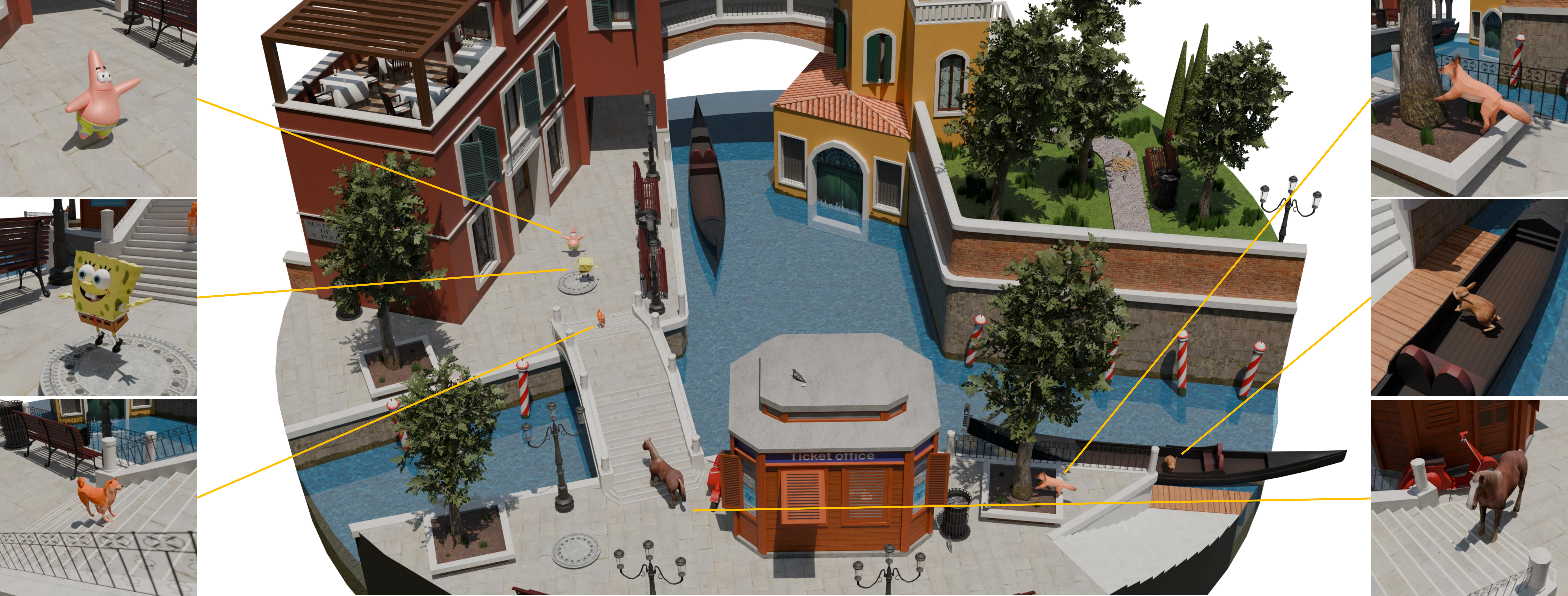}
    \vspace{-1.7em}
    \captionof{figure}{
        Given arbitrary scene and open-domain rigged objects, \textbf{\modelName} synthesizes articulation that respects the affordance and context.
        %
    } 
    \vspace{1.2em}
}]

\begin{abstract}
Rigged objects are commonly used in artist pipelines, as they can flexibly adapt to different scenes and postures.
However, articulating the rigs into realistic affordance-aware postures (\eg following the context, respecting the physics and the personalities of the object) remains time-consuming and heavily relies on human labor from experienced artists.
%
In this paper, we tackle the novel problem and design \modelName.
With a given context, such as the environment mesh and a text prompt of the desired posture, \modelName synthesizes articulation parameters for arbitrary and open-domain rigged objects obtained from the Internet. 
%
The task is incredibly challenging due to the lack of training data, and we do not make any topological assumptions about the open-domain rigs.
We propose using a 2D inpainting diffusion model and several control techniques to synthesize in-context affordance information.
Then, we develop an efficient bone correspondence alignment using a combination of differentiable rendering and semantic correspondence.
%
\modelName has stable convergence, completes in minutes, and synthesizes plausible affordance on different combinations of in-the-wild object rigs and scenes. 

\end{abstract}

\vspace{\secmargin}
\section{Introduction}
\label{sec:intro}
\vspace{\secmargin}

Recent advancements in generative models have enabled many applications in 3D content creation, such as synthesizing static objects~\cite{poole2023dreamfusion,liu2023zero,raj2023dreambooth3d,lin2023magic3d}, deformable objects~\cite{bah20244dfy,ling2024align}, and environments~\cite{hoellein2023text2room,shriram2024realmdreamer}.
As most research focuses on creating 3D assets, the ability to \textit{utilize} assets in real-world production (\eg gaming and artistic design) receives less attention.
In particular, we are interested in the workflow of articulating the rigged 3D assets commonly used in production for humans and animals.
Rigged objects are characterized by their reusability and flexibility in different environments compared to static assets.
However, placing these rigged objects into the scene relies heavily on human labor and remains challenging for experienced artists.
To ensure the resulting pose is visually plausible and both semantically and physically respects the scene geometry, it often takes artists hours to manipulate tens or hundreds of bones in the 3D space, which often also requires back-and-forth tweaking across the hierarchical topology of the rigs from different view directions.
These observations incentivize automating the process of placing articulated objects into the scene.
It reduces the labor for valuable artists and allows them to add final touches if necessary, compared to synthesizing the objects from scratch that produces finalized and unchangeable results.

The problem is non-trivial as there is no existing large-scale 3D object-scene composition data for model training, prohibiting intuitive solutions such as training a model to predict the feasible articulations for arbitrary rigged objects.
In real-world applications, the rigged objects and the environments are open sets with diverse appearances and complex semantics.
Their interaction produces a combinatorial explosion; training any prior model on small-scale data will suffer from the domain gap and lack practical value.
Moreover, the rigs' topology and joint placement do not have a universal definition.
Different artists can create different rigs for the same creature for various personal preferences or functionality needs.
For instance, rigs supporting facial expression and breathing motion will include extra bones.
It is non-trivial to distinguish which set of bones is necessary while posing the articulated object, and it is also challenging to design a model that works for arbitrary rigs that supply different functionalities. 

In this paper, we are interested in the problem of placing a rigged object into a scene with text prompt instructions, where the appearance and topology of the object and scene can be arbitrary.
The placement requires respecting the physical and semantic soundness of the object in the context.
This is often referred to as \textit{affordance} in robotics and computer vision.
The problem requires solving the per-bone SO$(3)$ transformations for linear blend skinning, which is a representation commonly used in all types of commercial rendering engines.
We propose an Affordance-Aware Articulation Synthesis framework called \modelName.
\modelName distills the object-scene interaction information from prompt-conditioned 2D inpainting models pre-trained on large-scale data.
As the diffusion model is pre-trained on large-scale open-set data to complete a part or the whole body of the objects with a given scene and text prompts, the model learns generalizable representations to produce in-context objects with plausible affordance.

Despite using a generative model to solve a generative task, challenges arise when applying a 2D foundation model to a 3D problem: efficiency and ambiguity.
For the efficiency problem, the commonly used Score Distillation Sampling (SDS)~\cite{poole2023dreamfusion} is known for its lengthy optimization process due to stochasticity and intensive computing due to the large number of sampling steps.
We argue that such inefficiency is especially unnecessary in our problem, as SDS expenses a significant part of computing on texture optimization.
At the same time, the gradients toward geometrical alternation are ineffective~\cite{zhang2024towards,zhu2024dreamhoi} and maintain a high variation.
Therefore, in \cref{sec:single-view}, we design an efficient geometry optimization paradigm using differentiable rendering and 2D-pixel correspondence.
The algorithm converges within a minute, compared to the hour-scale SDS requiring high-end GPUs.
For the ambiguity issue, as the 3D-to-2D projection loses the depth information, matching the appearance of a 3D articulated object with a 2D image reference is an ill-conditioned problem that has infinite results.
A natural solution is multi-view supervision, but existing multi-view diffusion models~\cite{liu2023syncdreamer,kong2024eschernet} fall short in our problem as they are object-centric and supply no affordance information.
In \cref{sec:multi-view}, we propose to employ a combination of partial denoising diffusion scheduling~\cite{haque2023instruct} along with grid prior~\cite{weber2024nerfiller}.
The former borrows contextual information (\eg current object pose, object appearance, and scene layout) that proposes slightly altered poses respecting the geometry from different angles.
The latter produces cross-view 3D consistent inpainting results by exchanging the information from various angles.

\issue{We summarize our main contributions as follow:}
\begin{itemize}[leftmargin=*]
    \item  We propose a novel and practical task in synthesizing affordance for open-domain rigged objects. To facilitate future research, we establish a benchmark dataset and code templates for converting rigged objects on the Internet to Python-programmable formats.
    \item  We address the challenging task with a data-free framework, supporting open-domain real-world rigged objects without topology or geometric assumptions.
    \item  Our framework features highly efficient articulation optimization with low variance. 
\end{itemize}

\begin{figure*}[t]
    \centering
    \includegraphics[width=\linewidth]{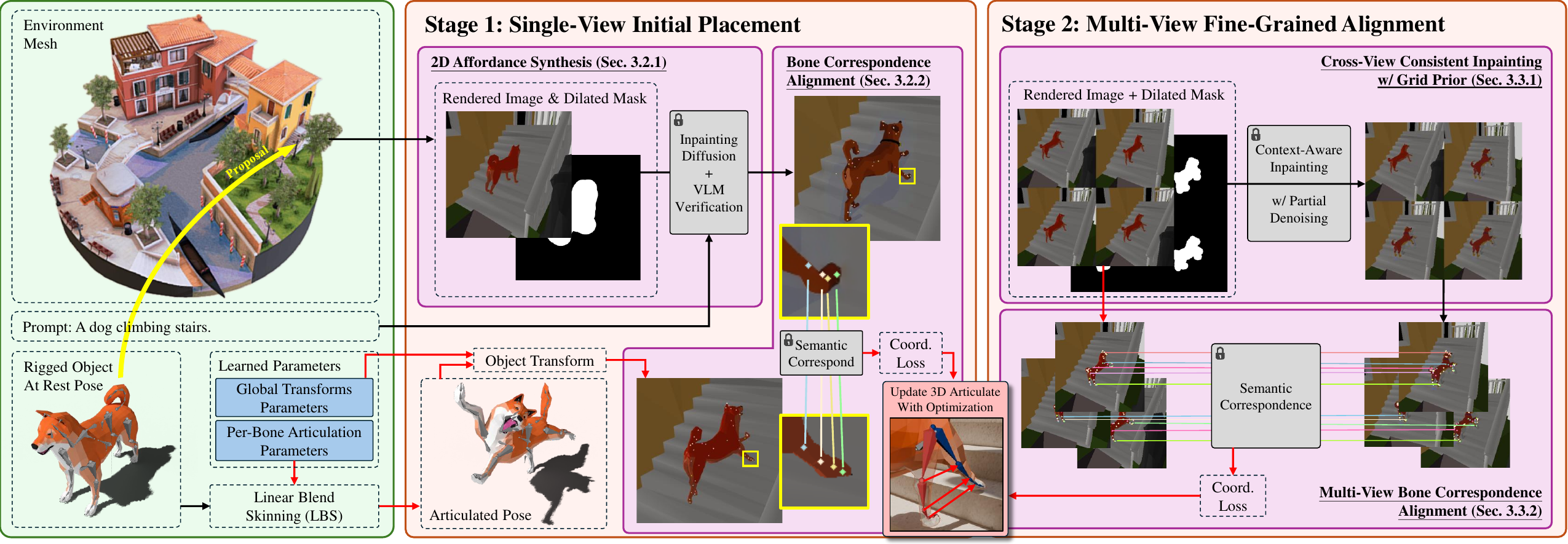}
    \vspace{-1.8em}
    \caption{
        \textbf{Overview.} 
        \textbf{(Left)} Our \modelName takes four inputs: The scene geometry, a rigged object, a text prompt describes the desired articulation, and an approximate location to perform the pose.
        The goal is to solve the object transformation and articulation parameters.
        \textbf{(Middle)} Our first stage aims to synthesize a course proposal posture, then optimizes the single-view pixel coordinate alignment with the current rest pose.
        The processing is fully and efficiently differentiable by using differentiable rendering and semantic correspondence.
        \textbf{(Right)} In the second stage, we use a combination of grid prior and partial denoising to synthesize cross-view consistent affordance reference, then optimizes the alignment in multiple views.
        In both stages, the optimization objective is equivalent to explicit 3D deformation, and we show such an optimization has a steady convergence.
    }
    \label{fig:pipeline}
    \vspace{-1em}
\end{figure*}
\vspace{\secmargin}
\section{Related Work}
\label{sec:related_work}
\vspace{\secmargin}

\issue{Affordance.}
Gibson~\cite{gibson1977theory} defines affordance as a property of the environment that allows the animal to interact with; as such, it implies the complementarity between the animal and the environment.
The early works~\cite{grabner2011makes,koppula2014physically,zhu2014reasoning,zhu2016inferring,fang2018demo2vec,nagarajan2019grounded,li2024one} thereby focus on analyzing the interaction between the human and the environment from images or videos.

\issue{Data-driven affordance synthesis.}
As the scale of available data increases, several works~\cite{wang2017binge,kulal2023putting} initiate the research on synthesizing the human affordance within environments in the 2D image space.
Later works~\cite{li2019putting,nagarajan2020learning,zhang2020generating,zhang2020place,hassan2021populating,wang2021synthesizing,wang2021scene,wang2022humanise,xiao2024unified,daiya2024collage} further expand the affordance research to 3D space.
These works mainly focus on training human affordance models for human-scene interaction with abundant established high-quality data.
Another line of work attempts to learn a non-human subject affordance model with limited training data, such as small-scale video data~\cite{cheng2023VirtualPets} and dense captures~\cite{yang2024ats}.
However, these works employ data-driven algorithms that require learning an affordance model from domain-specific data and lack generalization to open-domain subjects.

\issue{Data-free affordance synthesis.}
Despite the early success, the high-quality and annotated human-scene affordance data still needs to be improved.
Several recent works sought for zero-shot~\cite{li2024genzi,kimbeyond,xu2024interdreamer,zhu2024dreamhoi} settings that synthesize sophisticated human-scene or human-object interaction using pretrained foundation models.
However, all these approaches use human-centric prior models pretrained on large-scale human data, such as human pose estimation (in \cite{li2024genzi,kimbeyond,zhu2024dreamhoi}), human shape estimation (in \cite{kimbeyond}), and human-object affordance model (in \cite{xu2024interdreamer}).
They also leverage SMPL-X~\cite{pavlakos2019expressive}, which is meticulously extracted from thousands of 3D human scans.
Moreover, leveraging the diffusion model to solve human-related tasks is less challenging than other open-domain subjects, as large-scale datasets often include substantial human data.
Unfortunately, these assumptions are not available in our problem setting.
The subjects in our setting are not only open-domain, but the rigs' topology can be arbitrary with indefinite specifications.

\vspace{\secmargin}
\section{Method}
\label{sec:method}
\vspace{\secmargin}

\issue{Overview.}
We design \modelName compatible with the practical artist workflow, as it is a real-world challenge.
The inputs to \modelName{} include four components: a text prompt $\Gamma$, a scene mesh $\calS$, an approximal desired location $\mathbf{p} \in \mathbb{R}^3$ within $\calS$, and a rigged 3D object $\mathcal{C}$.
Our framework solves the object articulation parameters $\mathcal{A}$ (defines in \cref{sec:prelim}), and a set of global transformations $\mathcal{T}$ (includes translation $\in \mathbb{R}^3$, rotation $\in \mathbb{R}^{3 \times 3}$, and scaling factor $\in \mathbb{R}$).
In practice, we anticipate the users dragging and dropping the rigged object (at rest pose) to the approximate desired location and then providing a text-prompt description of the desired articulation.
Our algorithm will take over the remaining posing process and produce the plausible parameters $\mathcal{A}$ and $\mathcal{T}$.

In~\cref{fig:pipeline}, \modelName{} is designed in two stages.
First, in \cref{sec:single-view}, the single-view initial placement coarsely places the object from a single-view guidance.
Then, in \cref{sec:multi-view}, the multi-view fine-grained alignment stage finetunes the object's detailed posture from different angles, addressing the depth ambiguity from a single view.

\vspace{\subsecmargin}
\subsection{Preliminaries}
\label{sec:prelim}
\vspace{\subsecmargin}

\issue{Linear Blend Skinning (LBS) and articulation.}
We represent object articulation with linear blend skinning~\cite{lbs-course}, the most basic and commonly used formulation for direct skeletal shape deformation.
An articulated 3D object $\mathcal{C} = (V, B, W)$ is a collection of mesh vertices $V$, bones $B$, and skinning weights $W$.
The bones $B$ is a set of nodes organized in a hierarchical structure, representing the modifications to a parent node that will affect all its children.
Each bone is a control node that users can operate on. They typically control the relative position with the fixed-distance rotation.
The skinning weights $W \in \mathbb{R}^{|V| \times |B|}$ define the weighting factor of how a transformation in each bone will affect the location of all vertices.
As a displacement (\ie articulation) is applied to a bone, two things will happen: (a) the articulation also displaces all its children, and (b) all the vertices receive a weighted transformation based on the skinning weights.
In this work, we aim to solve the articulation parameters defined as a set of $\text{SO}(3)$ rotations for each bone, denoted as $\mathcal{A} \in \mathbb{R}^{|B| \times 3}$.

\issue{Semantic correspondence.}
In order to utilize the affordance information synthesized by the generative model, we use semantic correspondence~\cite{Zhang_2024_CVPR} to associate the pixel-level relationship between the rendered object and the synthesized image. 
Given two images $(I^s, I^t)$, semantic correspondence~\cite{ham2017proposal} was proposed to build the dense correspondence between the \textit{semantically} similar objects between the images that are not identical (\eg a rendered dog mesh and a realistic dog).
The semantic correspondence calculation is in two stages: feature extraction and similarity match.
Let $F = \mathbb{R}^{h \times w \times 3} \mapsto \mathbb{R}^{h \times w \times d}$ represent a pre-trained semantical feature extractor, mapping all pixel values of an input image $I$ into a $d$-dimensional feature space.
We denote $\ff_{u} = F(I)_{u}$, the semantic feature of a pixel located at $u$ in $I$. 
Then, for a query pixel in one image $I^s$, we denote its semantical feature as $\ff_q^s$.
Its semantic corresponding point $u^*$ in another image $I^t$ is determined by the pixel that maximizes the feature cosine similarity:
%
\begin{equation}
\label{eq:semantic_correspondence}
u^* = \arg \, \max_{u}  \, \mathop{\cos} \left( \ff_q^s, \ff_{u}^t\right) \, ,
\end{equation}
where $\cos(\cdot, \cdot)$ is the cosine similarity between two features.


\vspace{\subsecmargin}
\subsection{Single-View Coarse-Grained Placement}
\label{sec:single-view}
\vspace{\subsecmargin}

As motivated in \cref{sec:intro}, we design a training-free framework by leveraging the prompt-conditioned inpainting diffusion model as the primary source of affordance information.
The process has two steps: synthesizing the affordance and applying the distilled affordance to the rigged object.

\vspace{\subsubsecmargin}
\subsubsection{Affordance Synthesis With 2D Inpainting Model} 
\label{sec:first-stage-inpaint}
\vspace{\subsubsecmargin}

We start by determining a suitable viewpoint $k_0$ with the best visibility of $\mathcal{C}$. 
We sample multiple candidate cameras surrounding $\mathcal{C}$ at a fixed distance.
For each of the camera, $\mathcal{C}$ is rasterized to an RGB image $I_0$ and a silhouette mask $M_0$ with differentiable rasterization~\cite{pytorch3d}. 
%
In particular, we render two types of object silhouette masks, one with scene and one without scene.
The difference between these two masks calculates the occlusion rate.
We choose the camera with the most extensive silhouette and the lowest occlusion as the ideal viewpoint $k_0$.

Since $\calC$ is initially at its rest pose, its final pose after articulation will be different in position, size, rotation, and posture (\ie we solve $\mathcal{A}$ and $\mathcal{T}$).
We significantly dilate the silhouette mask by a radius $r$ to reserve the space for the inpainting model to generate different candidate poses, resulting in an updated mask $\hat{M}_0$.
The prompt-based inpainting diffusion model, $D$, then generates multiple candidate poses using different noises, written as $\hat{I}_0 = \text{VLM}( \, D(I_0, \hat{M}_0, \Gamma) \, )$.
%
%
Among the generated candidate inpainting results, we use a Vision Language Model (VLM)~\cite{openai2024chatgpt} to secure both the quality and whether the inpainted image aligns with the given text prompt. 
We provide the details of the procedure in Supplementary.


\vspace{\subsubsecmargin}
\subsubsection{Bone Correspondence}
\vspace{\subsubsecmargin}
\label{sec:method-bone-correspondence}

We design a novel \textit{bone correspondence} mechanism to transfer the affordance information from $\hat{I}_0$ to the current object posture parametrized by $\mathcal{A}$.
Previous works in human articulation synthesis~\cite{li2024genzi,kimbeyond,xu2024interdreamer,zhu2024dreamhoi} leverage pre-trained human models (\eg keypoint detection and pose estimation) to align the object pose.
However, there is no similar model for open-domain objects, mainly because the concept of keypoint is ill-defined in such a situation.
The challenging alignment problem between two semantically similar but different subjects is highly relevant to the semantic correspondence problem (covered in \cref{sec:prelim}).
We adopt the semantic correspondence framework to our scenario and design the bone correspondence.

During rasterizing the current object to image $I_0$, we identify the visible foreground vertices $V_\text{fg} \subseteq V$, where each $v_\text{fg} \in V_f$ corresponds to an image-space coordinate $u_\text{fg} \in U_\text{fg}$.
For each $v_\text{fg} \in V_\text{fg}$, we attribute the vertex to a bone $b_\text{fg} \in B$ that has the largest skinning weights $w_\text{fg}$, which implies the bone will contribute the most to the displacement of vertex during LBS.
Written as:
%
\begin{equation}
    \label{eq:bone_cluster}
    b_f = \mathop{\arg\max}_{b = \{1, \cdots, |B|\}} w_f \, .
\end{equation}
For each foreground coordinate $u_\text{fg}$, we identify the semantically corresponding coordinates $u^*_\text{fg}$ in $\hat{I}_0$ with \cref{eq:semantic_correspondence}.
%
Then, we average the semantic correspondence to obtain a more robust prediction, obtaining the $b$-th bone correspondence pairs $(\bar{u}_b, \bar{u}^*_b)$ with:
%
{\small
\begin{align}
    \bar{u}_b &= \frac{1}{N_\text{fg}^b} \,\, \sum_{v_\text{fg} \in V_\text{fg}}\mathbbm{1}\{b_\text{fg} = b\} \cdot u_\text{fg} \, , \label{eq:source_bone_center} \\
    \bar{u}^*_b &= \frac{1}{N_\text{fg}^b} \,\, \sum_{v_\text{fg} \in V_\text{fg}} \mathbbm{1}\{b_\text{fg} = b\} \cdot u^*_\text{fg} \, , \label{eq:target_bone_cetner}
\end{align}} \\ [-1em]
where $N^b_\text{fg}$ is the number of vertices belonging to $b$-th bone as noted in~\cref{eq:bone_cluster}. 
The bones that are invisible to the current view $k_0$ are ignored.
We filter outlier points with a feature cosine similarity score (in \cref{eq:semantic_correspondence}) lower than a threshold $\tau$ or outside a standard deviation.

Finally, we define the bone correspondence loss $\calL_{\text{BC}}$ using the computed bone correspondence between the rendered image $I_0$ and the inpainted image $\hat{I}_0$:
%
\begin{equation}
    \label{eq:bone_correspondence}
    \mathcal{L}_{\text{BC}} = \frac{1}{N_{\text{vis}}} \sum_{b = 1}^{|B|} \mathbbm{1}\{N^b_f > 0\} \cdot || \bar{u}_b - \bar{u}^*_b ||^2,
\end{equation}
where $N_{\text{vis}}$ denotes the number of bone whose $N_f^b > 0$.


\vspace{\subsubsecmargin}
\subsubsection{Bone Rotation Penalty}
\vspace{\subsubsecmargin}

As defined in \cref{sec:prelim}, the LBS bones are hierarchical, where the rotational angle of a leaf node inherits the rotation angles from all parents. 
Such a property makes the leaf nodes move faster than parent nodes (often near the mass center), overly compensating and overfitting the rotations that parent nodes should perform.
Such behavior often leads to unnatural limb angles, as shown in \cref{sec:ablation}.
To address this issue, we propose a hierarchical bone rotation penalty to regularize the rotation angle by bone hierarchy level.

Let $\ell_b$ be the hierarchy level of $b$-th bone in $B$ (the root node has $\ell_b=0$).
We define the hierarchical bone rotation penalty loss $\calL_{\text{RP}}$ as
%
\begin{equation}
\label{eq:rotation_penalization}
    \calL_{\text{RP}} = \frac{1}{|B|} \sum_{b \in B} \alpha^{\ell_b} \cdot ||\mathcal{A}_b||^2 \, ,
\end{equation}
where $\mathcal{A}_b$ is the articulation parameter of $b$, and $\alpha$ is a hyper-parameter re-weighting the penalty.
We find that weighting the loss with an exponential factor works well in practice.

\vspace{\subsubsecmargin}
\subsubsection{The First Stage Total Loss}
\vspace{\subsubsecmargin}
\label{sec:first-stage-total}

The total loss at this training stage is $\calL_{\text{SV}} = \lambda_{\text{BC}} \cdot \calL_{\text{BC}} + \lambda_{\text{RP}} \cdot \calL_{\text{RP}} + \lambda_{\text{SDF}} \cdot \calL_{\text{SDF}} \,$,
%
%
where the $\lambda$'s are hyper-parameters that weigh the losses. 
We use this objective to optimize the learnable parameters $\cal{A}$ and $\cal{T}$, the computational graph is end-to-end differentiable with differentiable rasterization. $\calL_{\text{SDF}}$ is a Signed Distance Field (SDF) loss that encourages physical contact and penalizes the object penetrating the scene. 
We first compute a voxel-based SDF $\Psi$ of the scene mesh, where each entry of the voxel records the distance to the mesh surface.
The SDF loss is
%
{\small
\begin{equation}
    \calL_{\text{SDF}} =
    \begin{cases}
        \displaystyle \,\, \mathop{\min}_{v \in V} \Psi(v) \,, & \text{if} \, \Psi(v) > 0 \, \forall v \in V \\
        \displaystyle \,\, \sum_{v \in V} \| \text{min}( \Psi(v), 0) \|_1 \,,  & \text{otherwise} \, , 
    \end{cases}
\end{equation}}
where the first term encourages object-scene contact, while the second term avoids penetration.

\vspace{\subsecmargin}
\subsection{Multi-View Fine-Grained Alignment}
\label{sec:multi-view}
\vspace{\subsecmargin}

The single-view alignment stage only approximates an initial pose.
Such a pose suffers from depth ambiguity, which may appear unnatural from a different viewpoint.
Although the ambiguity issue is intuitively solvable with multi-view guidance, maintaining cross-view consistency while inpainting multiple views is challenging.
We propose to overcome the problem with a combination of partial denoising and grid prior~\cite{weber2024nerfiller} to secure the cross-view consistency of the appearance and posture.

\vspace{\subsubsecmargin}
\subsubsection{Partial Denoising With Grid Prior.}
\vspace{\subsubsecmargin}

Starting with view selection, similar to the procedure in \cref{sec:first-stage-inpaint}, we scatter, rasterize, and compute the visibility of each camera.
At this stage, we only filter half of the cameras that have relatively low visibility and randomly sample four views from the remaining cameras, corresponding to images $\mathcal{I}=\{I_1, I_2, I_3, I_4\}$ and silhouette masks $\mathcal{M}=\{M_1, M_2, M_3, M_4\}$. 
As the masks are generated with rasterization from $\mathcal{C}$, the masks are strictly 3D consistent.
Following a similar procedure in the previous stage, we dilate the masks to reserve space for the model to synthesize appropriate articulation.

The challenges at this stage are twofold: cross-view consistency and keeping the inpainting model aware of the current object pose.
To address the former problem, inspired by \cite{weber2024nerfiller}, we utilize the grid prior to encourage the cross-view consistency, which spatially tiles $\mathcal{I}$ and $\mathcal{M}$ into a 2$\times$2 grid. 
Intuitively, the cross-view consistency is encouraged by sharing the context across tiled image features within the self-attention layers.
For the latter issue, we want to keep the inpainting model from synthesizing completely different poses, which destroys the initial results obtained from the previous stage. 
As a complete denoising diffusion process synthesizes entirely new content, we alternatively only execute a fraction of it, called partial denoising.
Let $T$ be the total denoising steps and $\gamma \in [0, 1]$ be the ratio of the denoising process we want to execute.
We first encode the tiled $\mathcal{I}$ back to the latent space with the encoder of the latent diffusion model. We add the noise at the denoising time step $T*\gamma$ (the noise level depends on the denoising scheduler) to recreate the noisy latent at the corresponding time step.
By completing the remaining denoising steps with the noisy latent, the tiled $\mathcal{M}$ and the text prompt $\Gamma$, we obtain the final inpainted images $\{\hat{I}_1, \cdots, \hat{I}_4\}$.
We show the effect of grid prior in \cref{sec:ablation}.

\vspace{\subsubsecmargin}
\subsubsection{Multi-view bone correspondence.}
\vspace{\subsubsecmargin}

Similar to \cref{sec:method-bone-correspondence}, we use bone correspondence loss to distill the affordance information from $\{\hat{I}_1, \cdots, \hat{I}_4\}$.
However, the cross-view consistency from the grid prior is approximated in the 2D space; it still sometimes remains contradictory in the actual 3D space when the sampled cameras are distant apart.
Therefore, in addition to simply computing a per-view bone correspondence loss, we improve the robustness of the loss with a loss threshold $\epsilon_t$ to exclude anomalous loss values.
The multi-view version of the bone correspondence loss $\calL_{\text{MVBC}}$ is
%
\begin{equation}
    \calL_{\text{MVBC}} = \frac{1}{N_{\text{valid}}} \sum_{m=1}^4 \mathbbm{1}\{\calL_\text{BC}^m < \epsilon_t \} \cdot \calL_\text{BC}^m,
\end{equation}
where $m$ is the index of the view, $\calL_\text{BC}^m$ is the single-view bone correspondence loss at view $k_m$, and $N_{\text{valid}}$ is the number of $\calL_\text{BC}^m$ lower than $\ell_t$.

\vspace{\subsubsecmargin}
\subsubsection{The Second Stage Total Loss}
\vspace{\subsubsecmargin}
Our multi-view refinement loss function is $\calL_{\text{MV}} = \lambda_{\text{MVBC}} \cdot \calL_{\text{MVBC}} + \lambda_{\text{RP}} \cdot \calL_{\text{RP}} + \lambda_{\text{SDF}} \cdot \calL_{\text{SDF}} \, $,
%
where the $\lambda$'s are loss weighting hyper-parameters.
Similar to the first stage, we optimize $\cal{A}$ and $\cal{T}$ with gradient descent.

\begin{figure*}[t]
    \centering
    \small
    \renewcommand{\tabcolsep}{0pt}
    \begin{tabular}{ccccccc}
        \begin{minipage}{0.141\linewidth}
            \centering
            \vspace{-7em}
            \includegraphics[width=.8\textwidth]{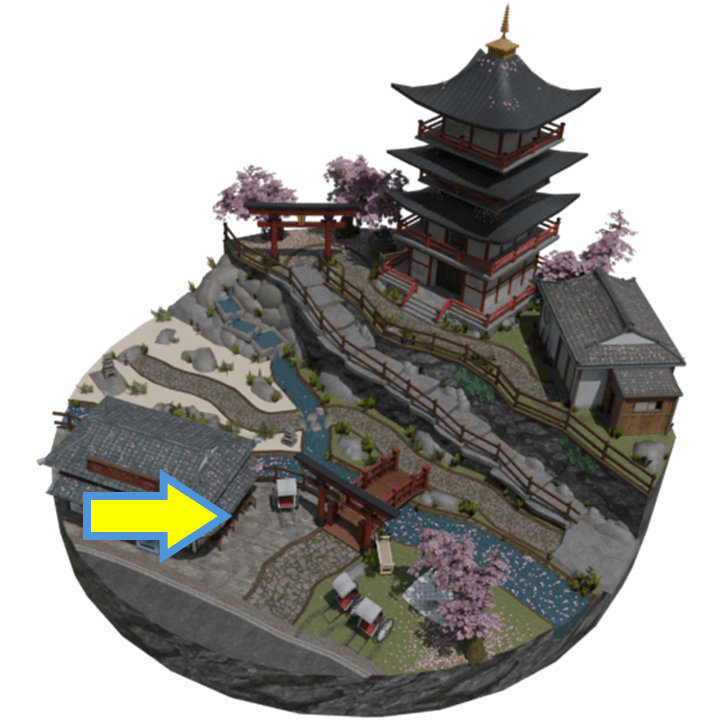} 
            \fontsize{6}{4}\selectfont
            Jumping down from a wooden chair to the ground.
        \end{minipage}
        \hfill&\hfill%
        \includegraphics[width=0.141\linewidth]{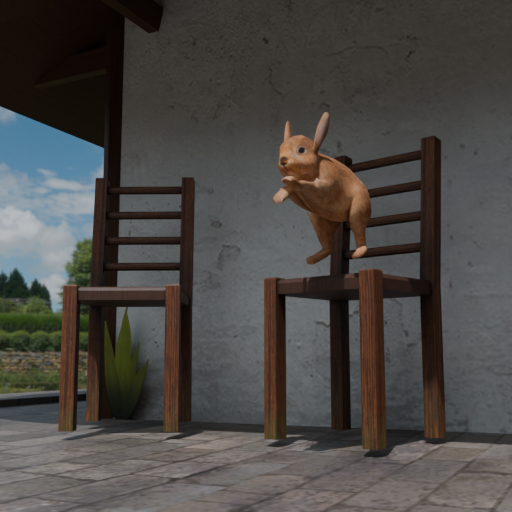} & 
        \includegraphics[width=0.141\linewidth]{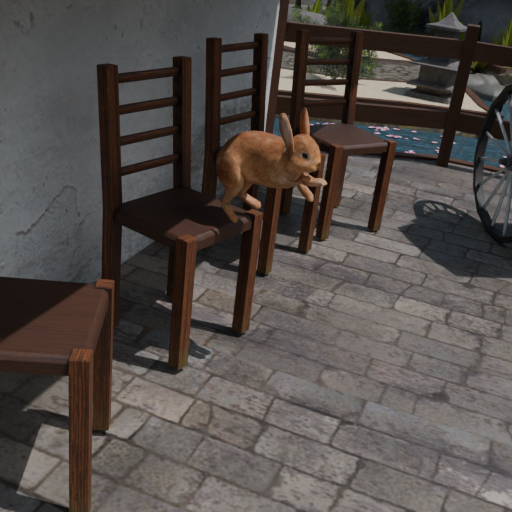}
        \hfill&\hfill%
        \includegraphics[width=0.141\linewidth]
        {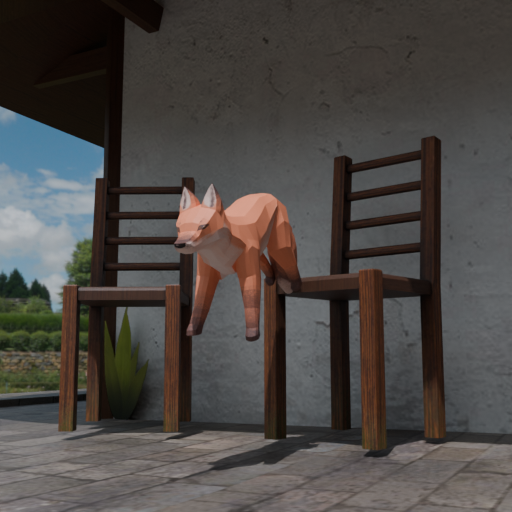} & 
        \includegraphics[width=0.141\linewidth]{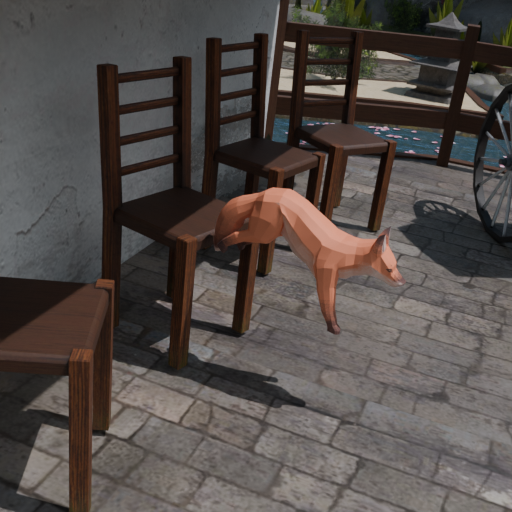}
        \hfill&\hfill%
        \includegraphics[width=0.141\linewidth]{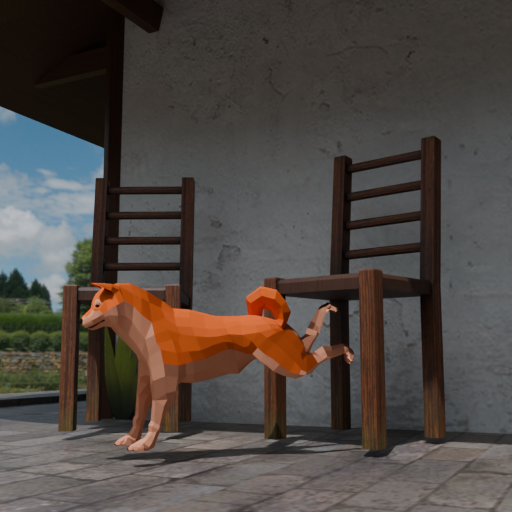} & 
        \includegraphics[width=0.141\linewidth]{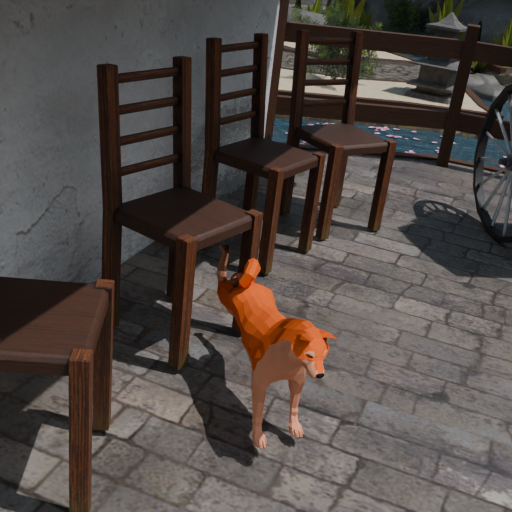} \\ [-.15em]
        \begin{minipage}{0.141\linewidth}
            \centering
            \vspace{-7em}
            \includegraphics[width=.8\textwidth]{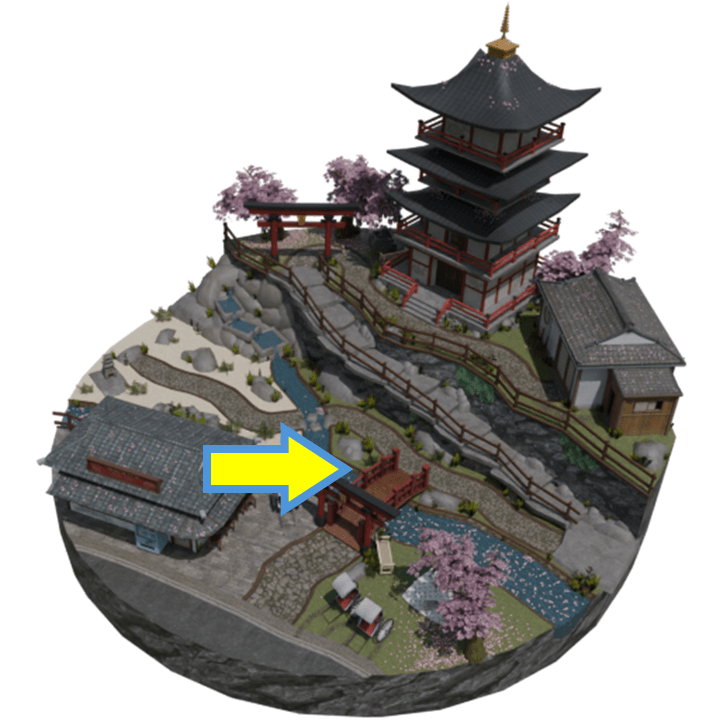} 
            \fontsize{6}{4}\selectfont
            Running on a \\ wooden bridge.
        \end{minipage}
        \hfill&\hfill%
        \includegraphics[width=0.141\linewidth]{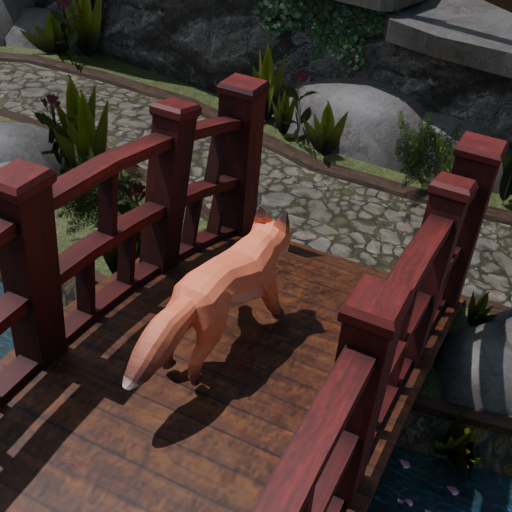} & 
        \includegraphics[width=0.141\linewidth]{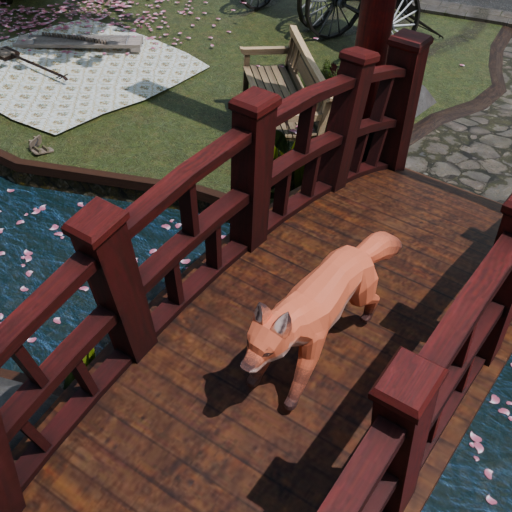} 
        \hfill&\hfill%
        \includegraphics[width=0.141\linewidth]{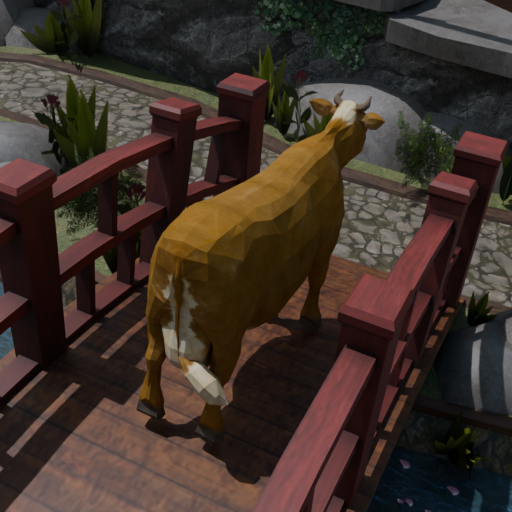} & 
        \includegraphics[width=0.141\linewidth]{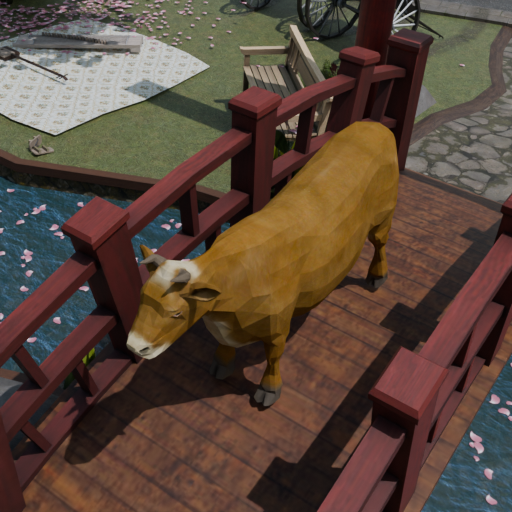} 
        \hfill&\hfill%
        \includegraphics[width=0.141\linewidth]{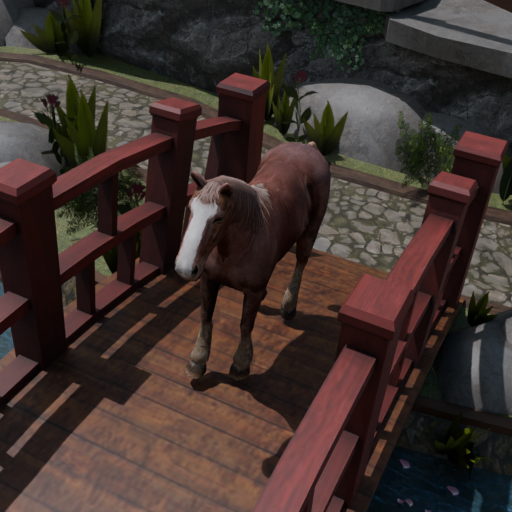} & 
        \includegraphics[width=0.141\linewidth]{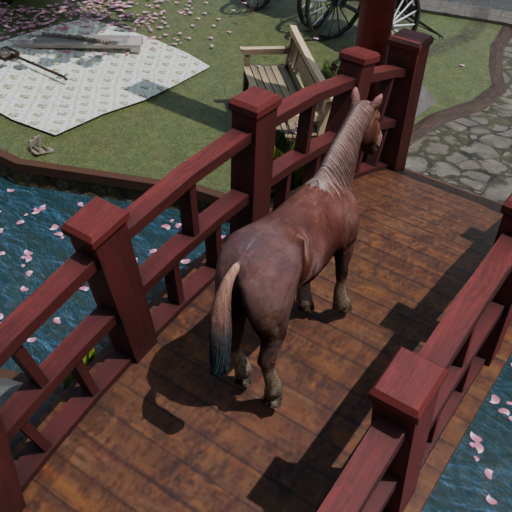} \\ [-.15em]
        \begin{minipage}{0.141\linewidth}
            \centering
            \vspace{-7em}
            \includegraphics[width=.8\textwidth]{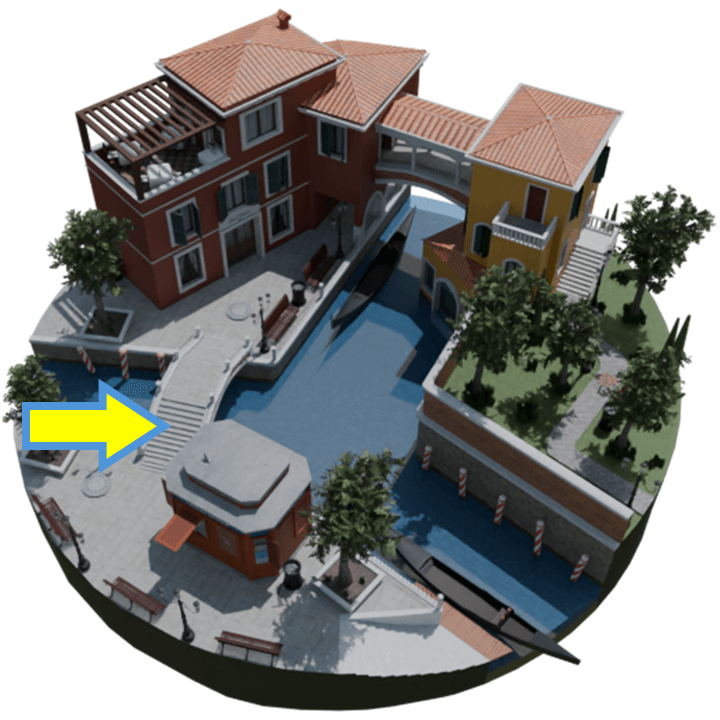} \\ [.3em] 
            \fontsize{6}{4}\selectfont
            Climbing stairs.
        \end{minipage} 
        \hfill&\hfill%
        \includegraphics[width=0.141\linewidth]{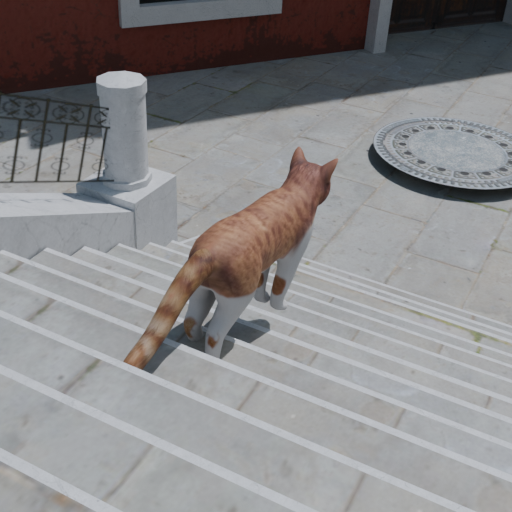} & 
        \includegraphics[width=0.141\linewidth]{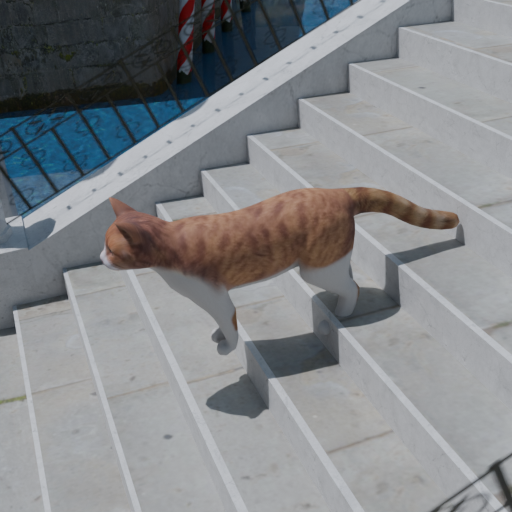} 
        \hfill&\hfill%
        \includegraphics[width=0.141\linewidth]{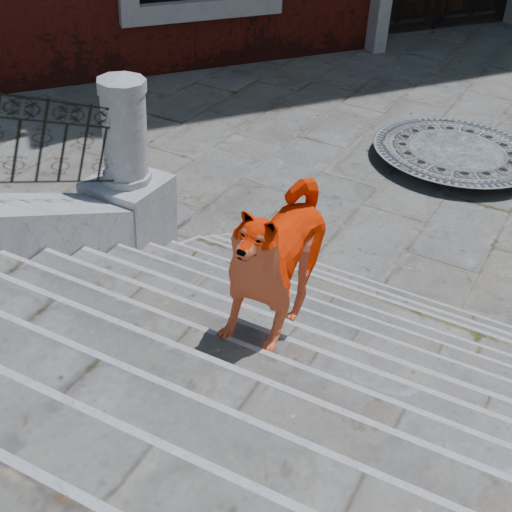} & 
        \includegraphics[width=0.141\linewidth]{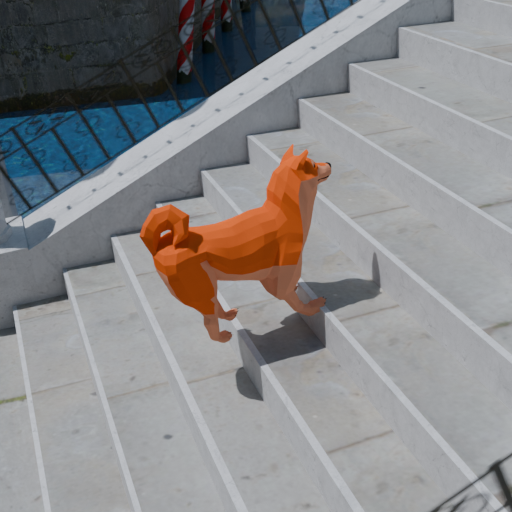} 
        \hfill&\hfill%
        \includegraphics[width=0.141\linewidth]{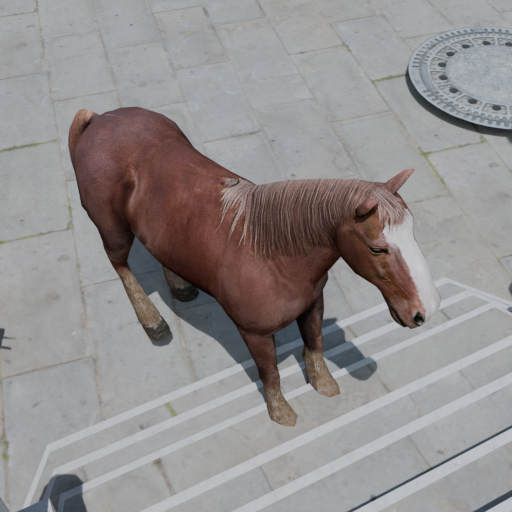} & 
        \includegraphics[width=0.141\linewidth]{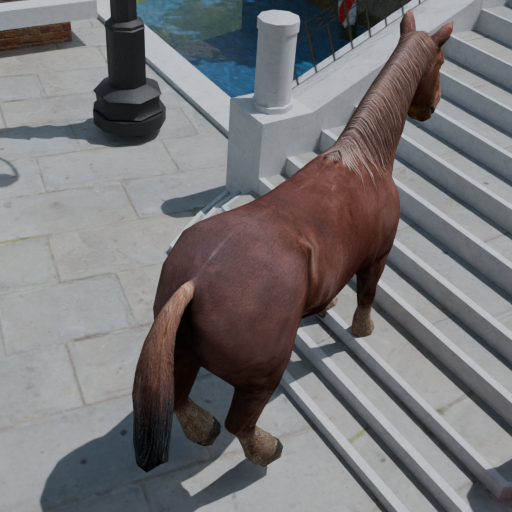} \\ [-.15em]
         \begin{minipage}{0.141\linewidth}
            \centering
            \vspace{-7em}
            \includegraphics[width=.8\textwidth]{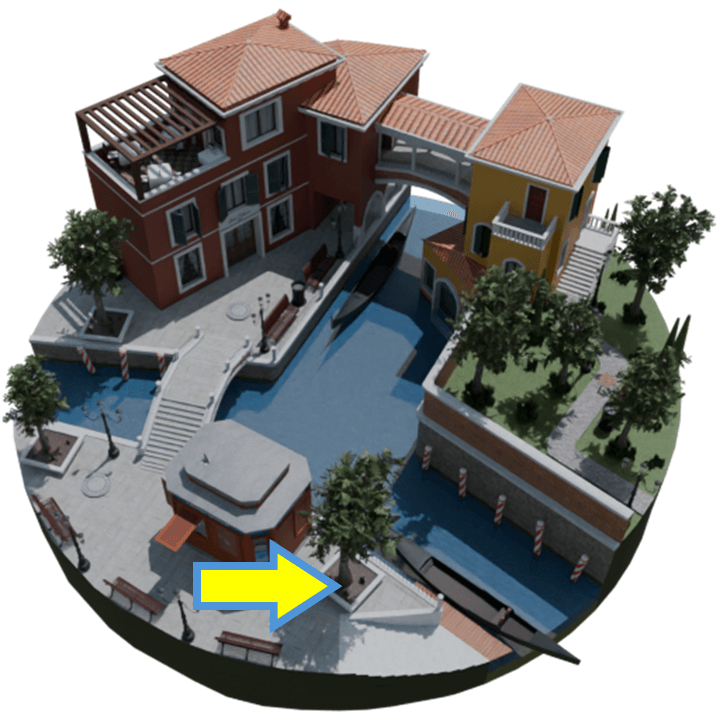} \\ [.3em]
            \fontsize{6}{4}\selectfont
            Climbing a tree.
        \end{minipage}  
        \hfill&\hfill%
        \includegraphics[width=0.141\linewidth]{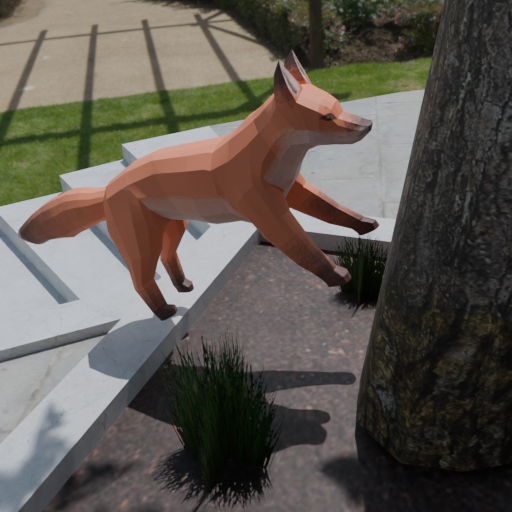} & 
        \includegraphics[width=0.141\linewidth]{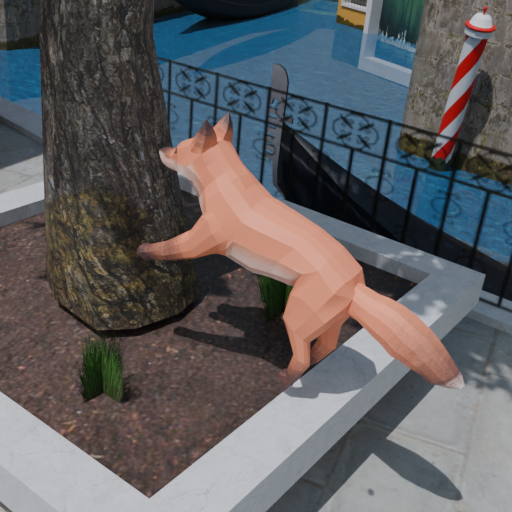} 
        \hfill&\hfill%
        \includegraphics[width=0.141\linewidth]{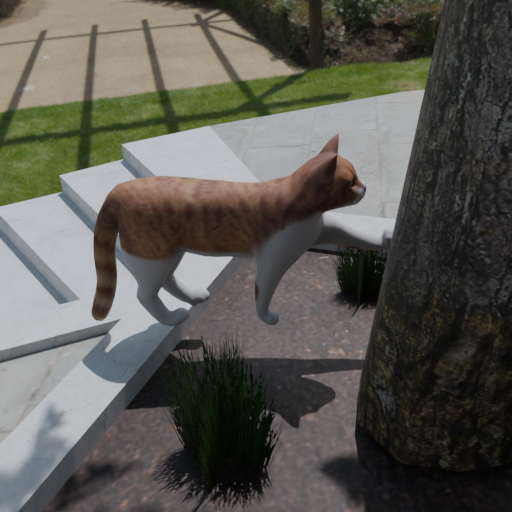} & 
        \includegraphics[width=0.141\linewidth]{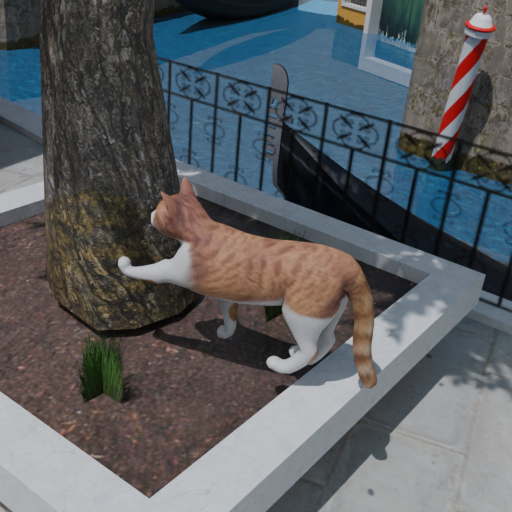} 
        \hfill&\hfill%
        \includegraphics[width=0.141\linewidth]{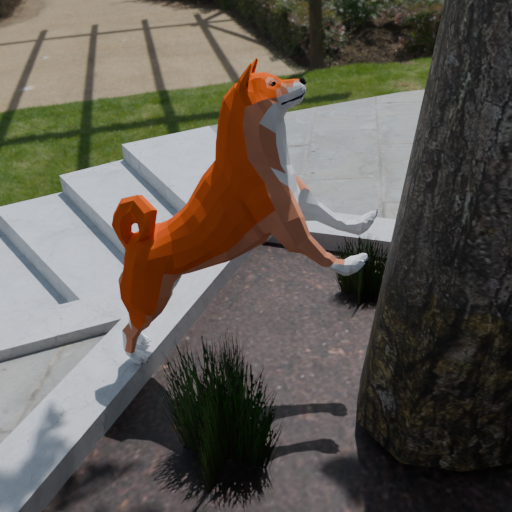} & 
        \includegraphics[width=0.141\linewidth]{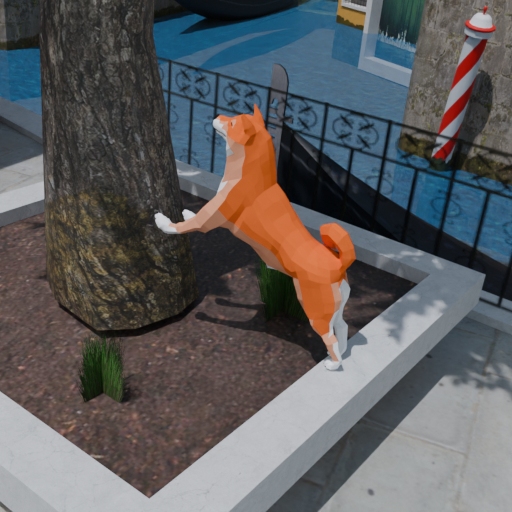} \\ [-.15em]
        Scene + Prompt & View 1 & View 2 & View 1 & View 2 & View 1 & View 2 \\
    \end{tabular}
    \vspace{-.7em}
    \caption{
        \textbf{The affordance-aware articulation synthesized with our \modelName.}
        For each scene-prompt-location composition, we use three different objects to show that our algorithm can adapt to arbitrary open-domain objects, maintain the physical soundness, and be aware of the object semantics (\eg the rabbit has a different jumping posture, the cat and dog has different tail signatures). Most importantly, the same object adapts to distinctive postures accord to different scene geometries, showing our results captures the nuance of affordance: \textit{the complementarity between the animal and the environment}~\cite{gibson1977theory}.
    }
    \label{fig:main-results-blender}
    \vspace{-1em}
\end{figure*}

\vspace{\subsubsecmargin}
\subsubsection{Iterative Refinement}
\vspace{\subsubsecmargin}
As we design the multi-view refinement stage aware of the current state of the object posture, we can iteratively apply the same process to refine the articulation from different views
To ensure the algorithm converges to a terminal state, we gradually reduce the synthesis freedom of the diffusion model (controlled by $\gamma$) at each iteration.
We found that running three rounds of iterative refinement is sufficient to reach a stable final state.
Such an iterative refinement is entirely optional.
In \cref{fig:exp-sds-convergence}, we show the first iteration of multi-view alignment has nearly reached the terminal state.

\vspace{\secmargin}
\section{Experiments}
\label{sec:experiment}
\vspace{\secmargin}

\issue{Datasets.}
We collect six rigged objects from the Internet, including \textit{shiba inu, fox, horse, cow, cat and rabbit}. We establish code templates and pipelines to convert rigged objects obtained from the Internet to python-programmable formats. 
The pipeline first extracts the bone hierarchy, skinning weights, and bone constraints from artist designed rigged objects.
Then, we convert the format compatible to LBS implemented in PyTorch~\cite{paszke2019pytorch}, and verify the correctness of the conversion.
%
Note that the effort is substantial, as a lot of the objects may fail the verification due to non-standard rigging or weight painting by amateur artists.
To facilitate future effort on collecting more rigged object data, we will make our pipeline publicly available. 
%
We test these objects on 3D scenes released by~\cite{li2024genzi}. 
Combining with text prompts varying at different positions, we gather a total of 20 object-scene interaction pairs for evaluation.

\issue{Metrics.}
To evaluate the semantic alignment and physical plausibility of the results, we adopt an evaluation protocol similar to the human-scene interaction task~\cite{li2024genzi}.
%
\begin{itemize}[leftmargin=*]
    \item \textbf{CLIP score~\cite{hessel2021clipscore}} measures the semantic alignment between the rendered images and prompts by comparing the feature similarity between the image and prompt features extracted with a pre-trained CLIP ViT-B/32 model~\cite{radford2021learning}. We sample $k$ viewpoints for each unique prompt-object-scene composition and calculate the average.
    \item \textbf{Non-collision score}~\cite{zhang2020generating,zhao2022compositional,li2024genzi} measures the ratio of vertices with positive SDF values from $\Psi$ (\cref{sec:first-stage-total}), evaluating whether the object penetrates the scene mesh.
    \item \textbf{Contact score}~\cite{zhang2020generating,zhao2022compositional,li2024genzi} measures the ratio of object-scene pair where the object contacts with the scene mesh. The contact is evaluated by whether any of the vertex has a non-positive SDF value measured with $\Psi$.
\end{itemize}
%
%


\issue{Baseline method.} 
As there is no previous work directly solving our task, we implement an SDS baseline to update the parameters. 
We use the publicly available StableDreamFusion~\cite{stable-dreamfusion} implementation.
To stabilize the training and stochasticity, we use HiFA~\cite{zhu2024hifa} to schedule the noise level from high to low base on the number of iterations into the optimization.
We create cameras at five different height levels (\ie 10, 25, 40, 55 and 70 degrees), with 20 equally-spacing cameras at each level pointing toward the object at the same distance, and filter out cameras that has less than 80\% visibility.
The selection process typically keeps 70 to 90 cameras in the end.

\issue{Hyperparameters.} We report more implementation details in Supplementary.

\vspace{\subsecmargin}
\subsection{Qualitative Evaluation}
\vspace{\subsecmargin}

\issue{Main results.}
\cref{fig:main-results-blender} presents the results of our proposed method across four distinct scenes, each guided by a unique text prompt to drive the object’s interaction with the scene. 
For each scene/prompt pair, we show three different objects performing a similar postures specified by the same prompt.
The results show that our method can adapt to different object-scene compositions, and aware of the semantic-level affordance of different objects (\eg the rabbit jumps differently, the tails are in different posture for dogs and cats).
%
We render the results in different views, showing the positioning is 3D correct without severe counter-physic behaviors.
Our method consistently produces realistic articulations with reasonable affordance across different views, verifying the multi-view alignment successfully demystifies the depth ambiguity.
%
%

\begin{figure}[t]
    \centering
    \renewcommand{\tabcolsep}{1pt}
    \renewcommand{\arraystretch}{0.5} 
    \begin{tabular}{lcccc}
         \parbox[c]{.8em}{\rotatebox[origin=c]{90}{\small Rest Pose\hspace{-5em}}} &
         \includegraphics[width=0.23\linewidth,trim={1cm 4cm 5cm 2cm},clip]{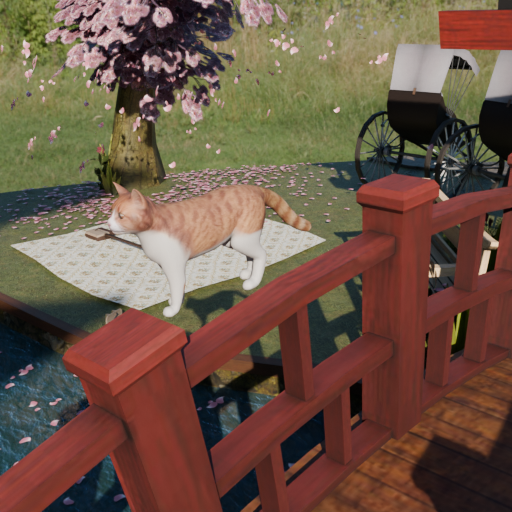} &
         \includegraphics[width=0.23\linewidth,trim={1cm 3cm 2cm 0cm},clip]{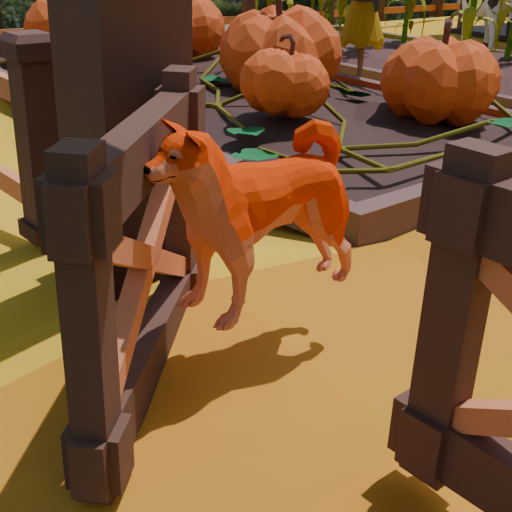} &
         \includegraphics[width=0.23\linewidth,trim={2cm 1cm 1cm 2cm},clip]{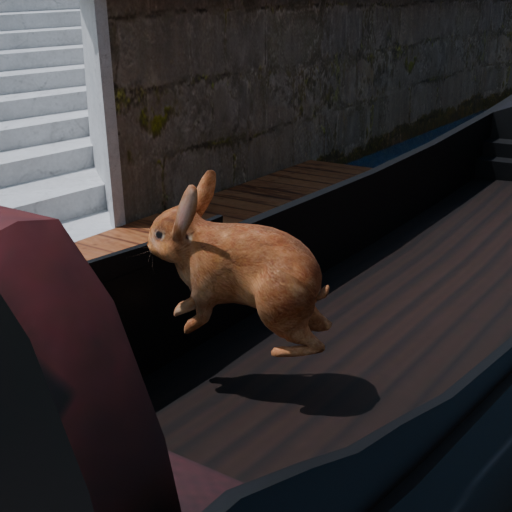} &
         \includegraphics[width=0.23\linewidth,trim={4cm 4cm 0cm 0cm},clip]{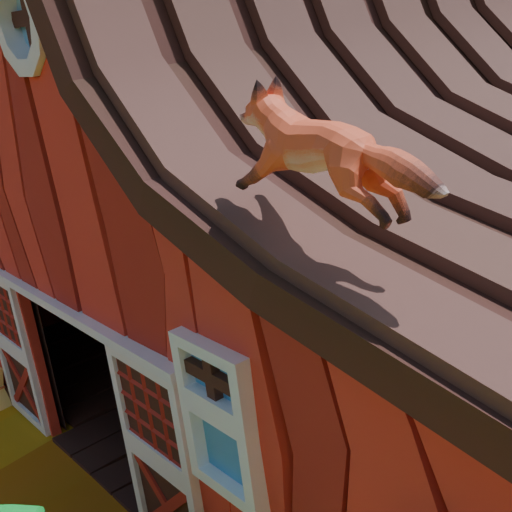} \\
         \parbox[c]{.8em}{\rotatebox[origin=c]{90}{\small SDS\hspace{-5em}}} &
         \includegraphics[width=0.23\linewidth,trim={1cm 4cm 5cm 2cm},clip]{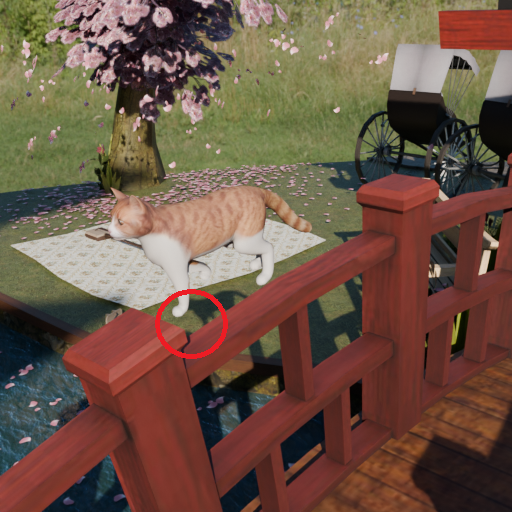} &
         \includegraphics[width=0.23\linewidth,trim={1cm 3cm 2cm 0cm},clip]{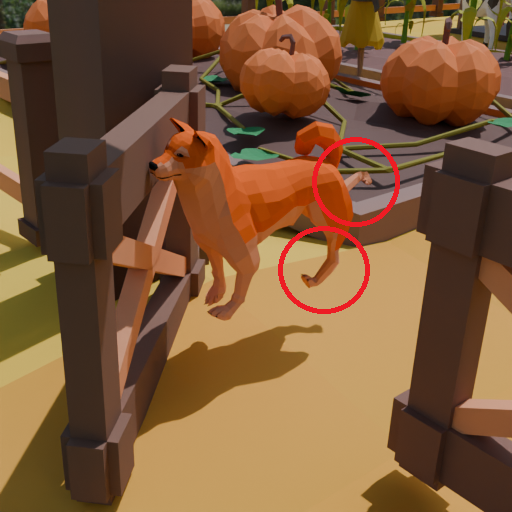} &
         \includegraphics[width=0.23\linewidth,trim={2cm 1cm 1cm 2cm},clip]{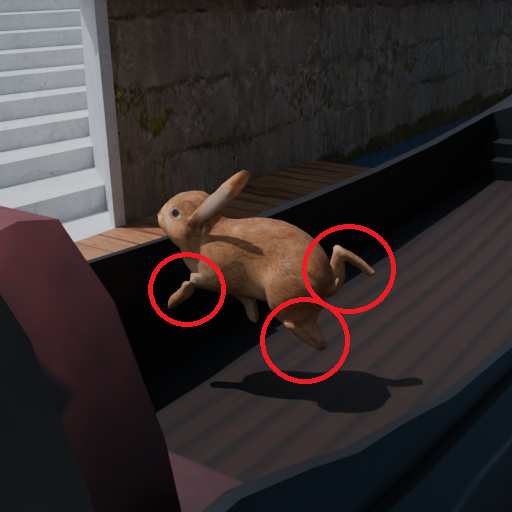} &
         \includegraphics[width=0.23\linewidth,trim={4cm 4cm 0cm 0cm},clip]{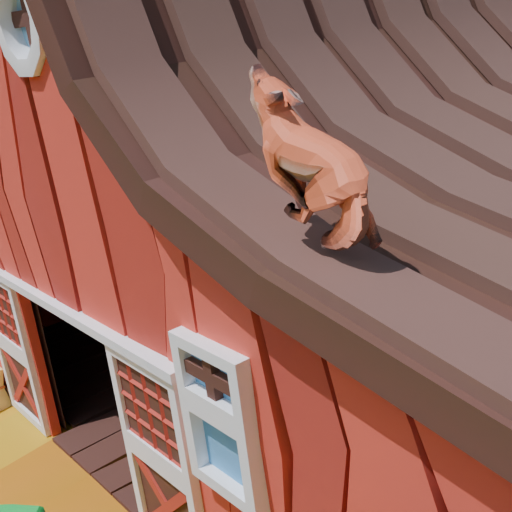} \\
         \parbox[c]{.8em}{\rotatebox[origin=c]{90}{\small Ours\hspace{-5em}}} &
         \includegraphics[width=0.23\linewidth,trim={1cm 4cm 5cm 2cm},clip]{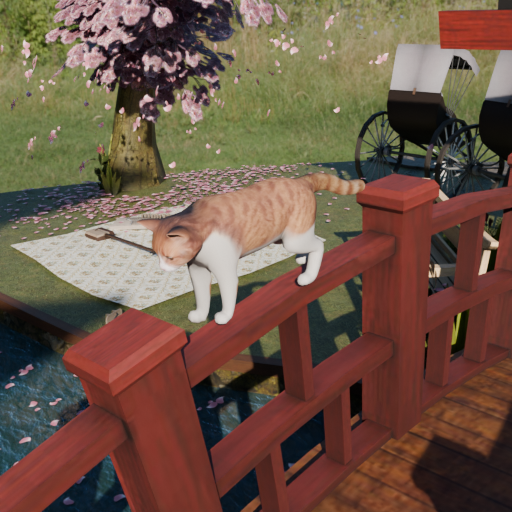} &
         \includegraphics[width=0.23\linewidth,trim={1cm 3cm 2cm 0cm},clip]{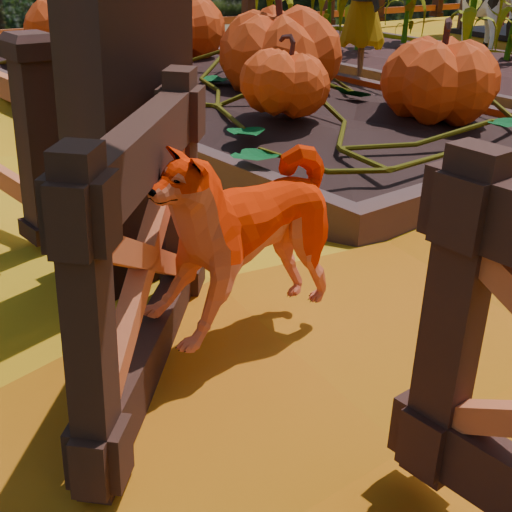} &
         \includegraphics[width=0.23\linewidth,trim={2cm 1cm 1cm 2cm},clip]{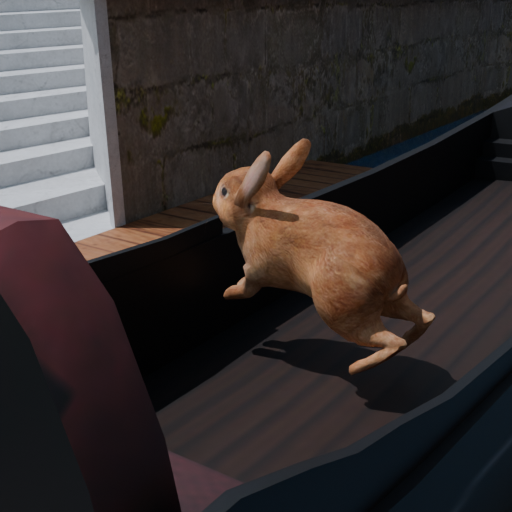} &
         \includegraphics[width=0.23\linewidth,trim={4cm 4cm 0cm 0cm},clip]{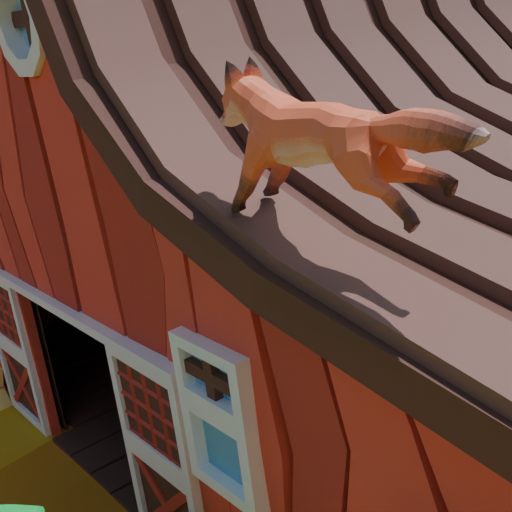} \\
         &
         {\fontsize{6}{4}\selectfont Running on railing.} &
         {\fontsize{6}{4}\selectfont Pushing the fence.} & 
         {\fontsize{6}{4}\selectfont Jumping from boat.} & 
         {\fontsize{6}{4}\selectfont Running on rooftop.} \\
    \end{tabular}
    \vspace{-.7em}
    \caption{
        \textbf{Comparisons.}
        SDS has a limited pose change from the rest pose, or creates unnaturally distorted limbs (\eg the legs of the shiba inu and rabbit). 
        Our method produces more natural posture, while the added articulation better resembles the affordance.
    }
    \label{fig:comparison-sds}
\end{figure}

\issue{Comparison with SDS.}
\cref{fig:comparison-sds} show the qualitative comparison with the baseline method implemented by SDS. All methods use similar hyperparameters. As previous study~\cite{zhang2024towards,zhu2024dreamhoi} has shown that SDS is ineffective in updating object transformation, we similarly observe that SDS is not effective in updating the articulation parameters.
Moreover, the gradient from SDS tend to create large spikes in optimizing the object scale, forcing us to manually set the learning rate of global transformations to $10{,}000$ times smaller than the articulation learning rate.
On the other hand, our method has stable convergence in all parameters, steadily follow the affordance information supplied by the generative model, and leading to more plausible final articulation.

%



\vspace{\subsecmargin}
\subsection{Quantitative Evaluation}
\vspace{\subsecmargin}

\begin{table}[t]
    
    \centering
    \small
    \renewcommand{\tabcolsep}{3pt}
    \begin{tabular}{lccc}
        \toprule
        Method & CLIP score ($\uparrow$)   & Non-Collision ($\uparrow$)  & Contact ($\uparrow$) \\
        \midrule
        Initial & 0.291 & - & - \\
        SDS     & 0.290 & 0.972 & 0.529\\
        \modelName (ours) & \textbf{0.297} & \textbf{0.993} & \textbf{0.6} \\
        \bottomrule
    \end{tabular}
    \vspace{-.5em}
    \caption{
    \textbf{Semantic correctness and physical plausibility.} 
    \modelName has the best performance in all metrics. 
    \textit{\textbf{Initial}} is the object rest pose.
    CLIP score evaluates the semantic consistency; the non-collision and contact loss measures the physical correctness between the object and scene.
    All metrics are the higher the better.
    %
    }
    \label{tab:quantitative_evaluation}
    
\end{table}
\begin{figure}[t]
    \centering
    \renewcommand{\tabcolsep}{1pt}
    \renewcommand{\arraystretch}{0.5} 
    \begin{tabular}{ccc}
         \includegraphics[width=.27\linewidth,trim={4cm 8cm 5cm 2.5cm},clip]{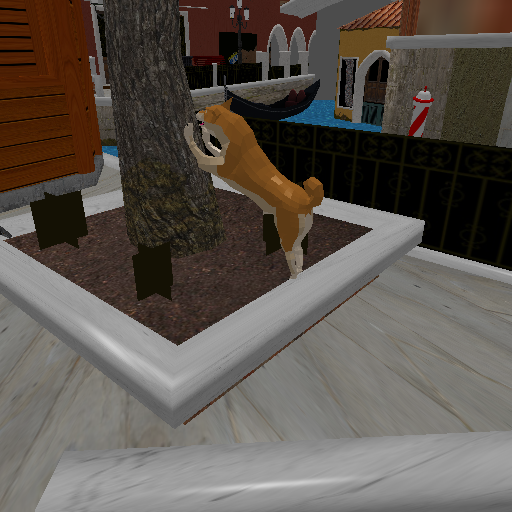} &
         \includegraphics[width=.27\linewidth,trim={4cm 8cm 5cm 2.5cm},clip]{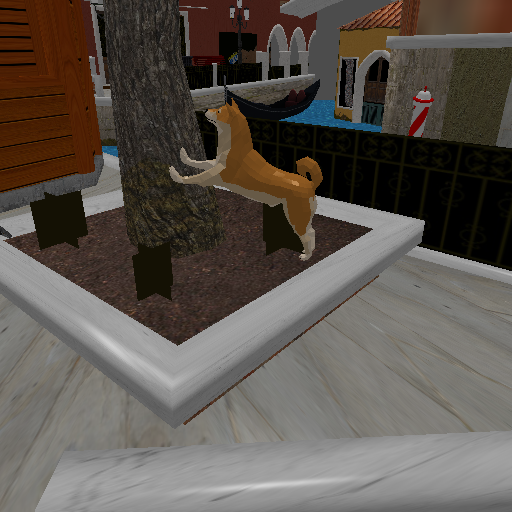} &
         \includegraphics[width=.27\linewidth,trim={4cm 8cm 5cm 2.5cm},clip]{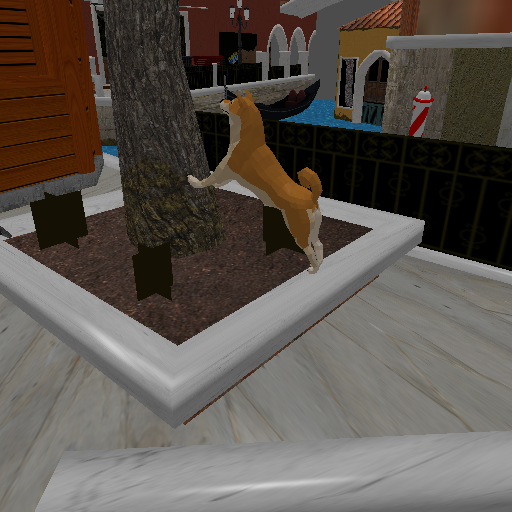} \\
         \includegraphics[width=.27\linewidth,trim={4cm 8cm 5cm 2.5cm},clip]{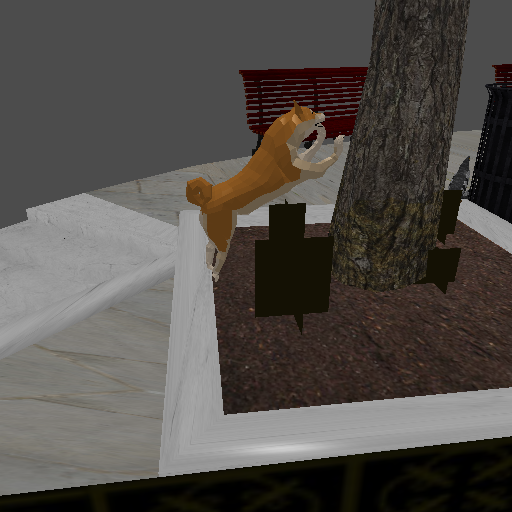} &
         \includegraphics[width=.27\linewidth,trim={4cm 8cm 5cm 2.5cm},clip]{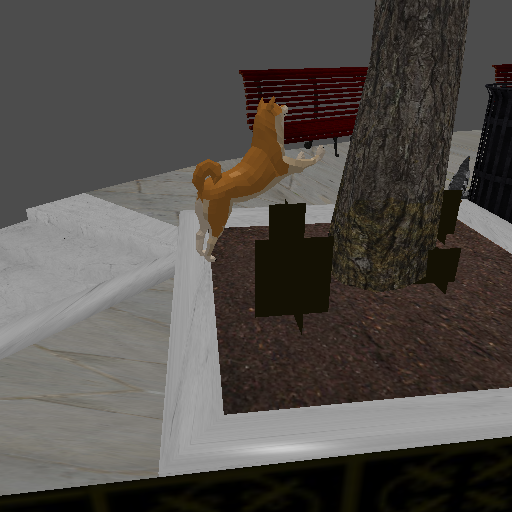} &
         \includegraphics[width=.27\linewidth,trim={4cm 8cm 5cm 2.5cm},clip]{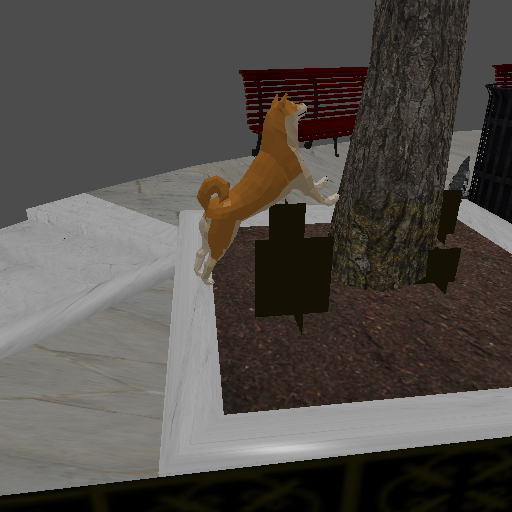} \\
         {\fontsize{6}{4}\selectfont w/o BR Penalty} &
         {\fontsize{6}{4}\selectfont w/o MV Alignment} &
         {\fontsize{6}{4}\selectfont Full (ours)} \\
    \end{tabular}
    \vspace{-.5em}


    \caption{
        \textbf{Ablation study.} 
        We show a sample of \textit{dog attempting to climb tree} from two views. 
        Removing bone rotation penalty (BR) causes unnatural limb bending, while omitting our second stage multi-view alignment (MV) leads to floating due to single-view depth ambiguity.
        Combining all methods lead to the best posture.
    }
    \label{fig:ablation}
\end{figure}

\begin{figure*}
    \centering
    %
    %
    %
    %
    \hfill
    \begin{subfigure}[t]{0.3\linewidth}
        \includegraphics[width=\textwidth]{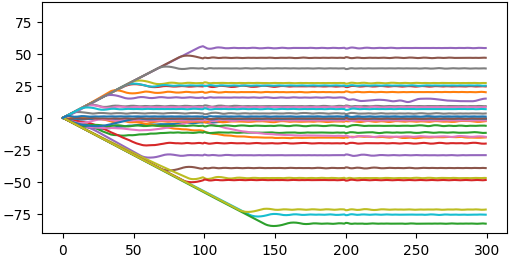}
        \caption{Ours (no learning rate decay)}
    \end{subfigure}
    \hfill
    \begin{subfigure}[t]{0.3\linewidth}
        \includegraphics[width=\textwidth]{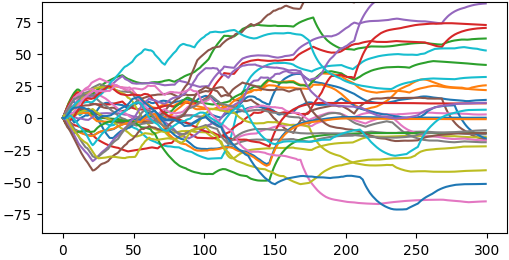}
        \caption{SDS w/o learning rate decay}
    \end{subfigure}
    \hfill
    \begin{subfigure}[t]{0.3\linewidth}
        \includegraphics[width=\textwidth]{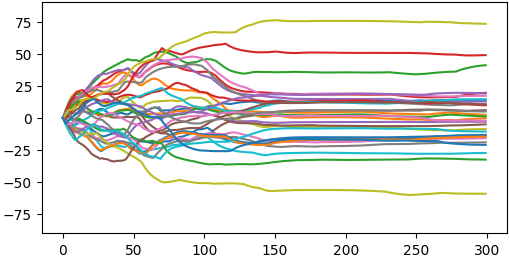}
        \caption{SDS w/ learning rate decay}
    \end{subfigure}
    \hfill
    \vspace{-.7em}
    \caption{
        \textbf{Our approach provides steady convergence.}
        We visualize the bone rotation in degrees (y axis) to optimization iterations (x axis), each line represents a unique bone.
        All methods use similar hyperparameters.
        Our approach (no learning rate decay) has a clear converge direction.
        In contrast, SDS does not have a consistent converge target, even with HiFA scheduling~\cite{zhu2024hifa} that sets a low noise rate by the end of optimization. 
        Adding learning rate scheduling mitigates the issue (still unstable at end), but restricts the change in angle.
    }
    \label{fig:exp-sds-convergence}
\end{figure*}
\begin{figure}[t]
    \centering
    \renewcommand{\tabcolsep}{1pt}
    \renewcommand{\arraystretch}{0.5} 
    \begin{tabular}{lcccc}
        \parbox[c]{.8em}{\rotatebox[origin=c]{90}{\fontsize{8}{4}\selectfont  w/o grid prior\hspace{-6em}}} &
        \includegraphics[width=.23\linewidth]{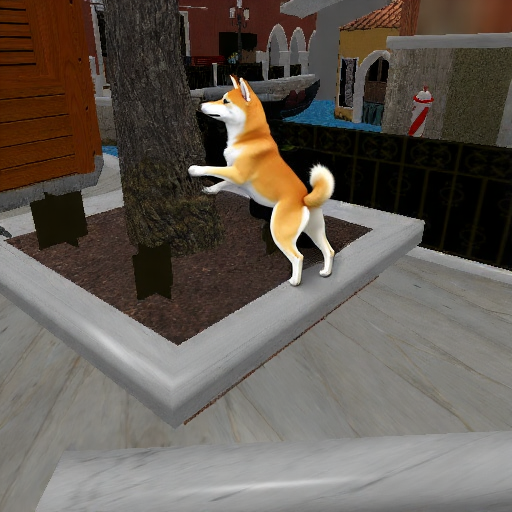} &
        \includegraphics[width=.23\linewidth]{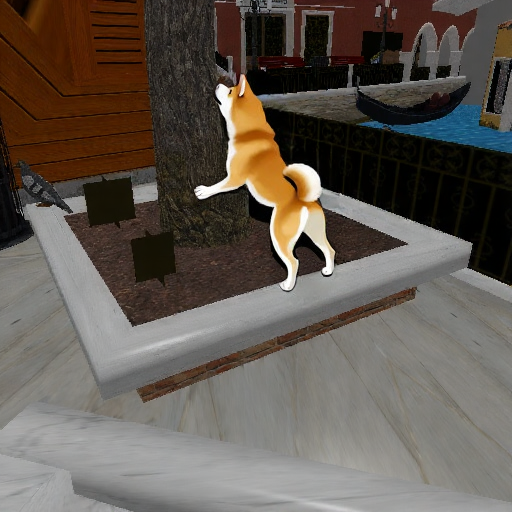} &
        \includegraphics[width=.23\linewidth]{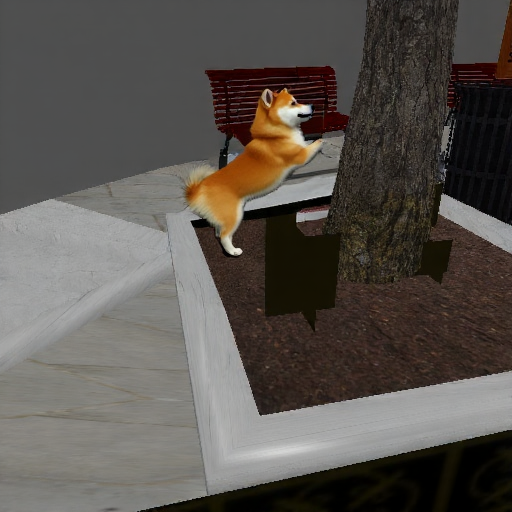} &
        \includegraphics[width=.23\linewidth]{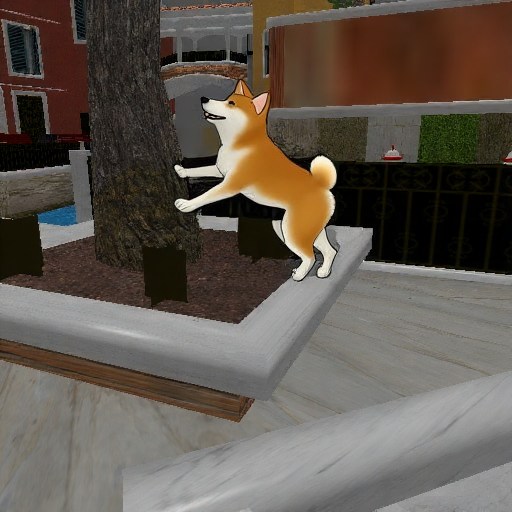} \\
        \parbox[c]{.8em}{\rotatebox[origin=c]{90}{\fontsize{8}{4}\selectfont  w/ grid prior \hspace{-6em}}} &
        \includegraphics[width=.23\linewidth]{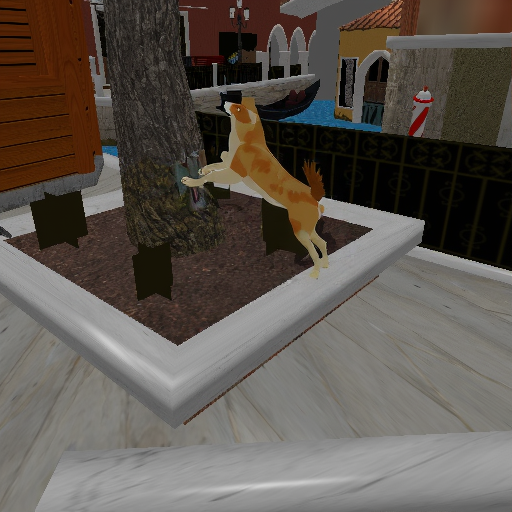} &
        \includegraphics[width=.23\linewidth]{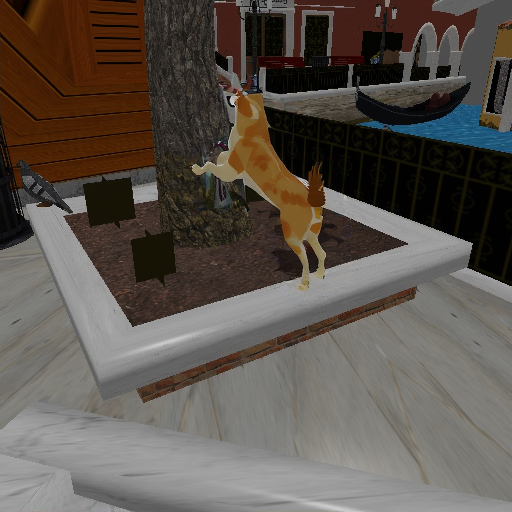} &
        \includegraphics[width=.23\linewidth]{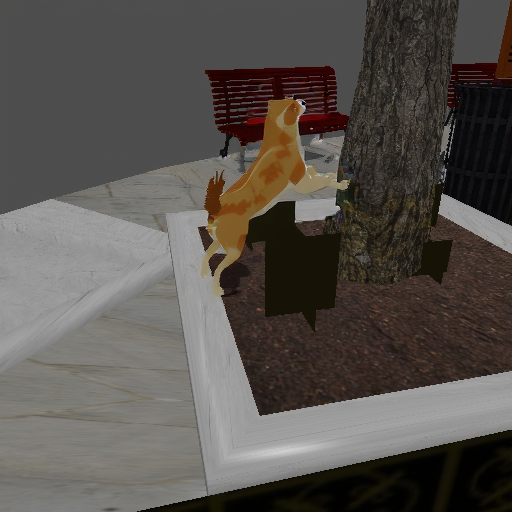} &
        \includegraphics[width=.23\linewidth]{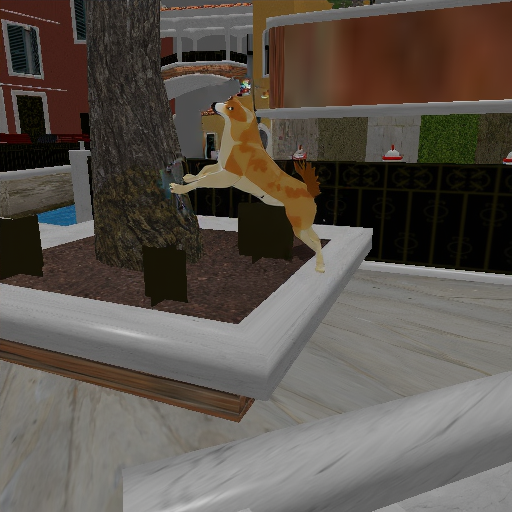} \\
    \end{tabular}
    \vspace{-.7em}
    \caption{
        \textbf{Comparing inpainting strategies.}
        Our strategy with grid prior maintains better cross-view consistency, compared to individually inpaint each view produces inconsistent limb locations.
    }
    
    \label{fig:grid_prior}
\end{figure}
\begin{figure}[t]
    \centering
    \renewcommand{\tabcolsep}{1pt}
    \renewcommand{\arraystretch}{0.5} 
    \begin{tabular}{ccc}
        \includegraphics[width=0.3\linewidth]{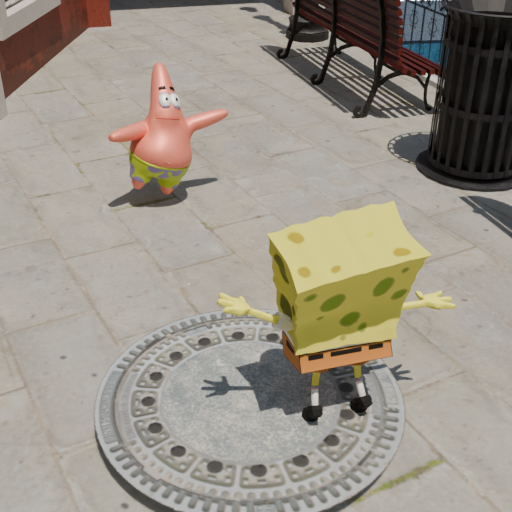} &
        \includegraphics[width=0.3\linewidth]{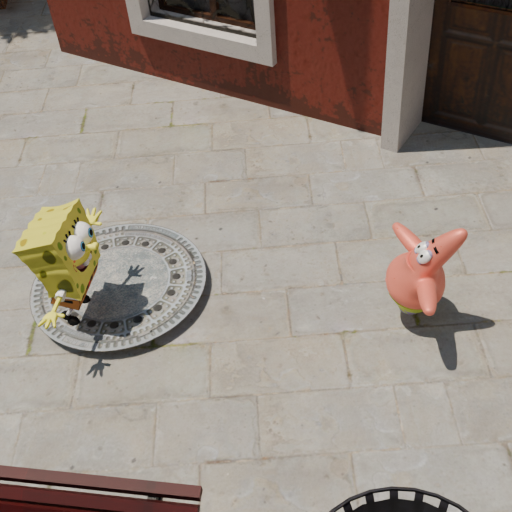} &
        \includegraphics[width=0.3\linewidth]{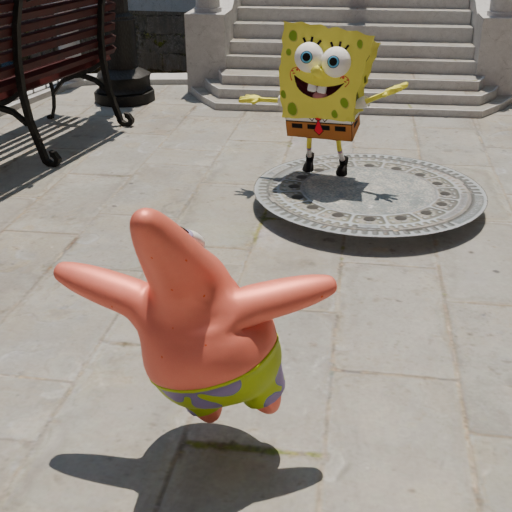} \\
        {\fontsize{8}{4}\selectfont View 1} &
        {\fontsize{8}{4}\selectfont View 2} &
        {\fontsize{8}{4}\selectfont View 3} \\
    \end{tabular}
    \vspace{-.7em}
    \caption{
        \textbf{Additional out-of-domain objects.} 
        \modelName can generalize to cartoon characters with unusual appearance and geometry.
    }
    \label{fig:out-of-domain}
    \vspace{-1em}
\end{figure}

\cref{tab:quantitative_evaluation} presents the quantitative evaluation of our method compared to initial placement (\ie positioning the object in a rest pose within the scene) and the SDS baseline. Our method consistently outperforms both initial placement and SDS in terms of semantic accuracy and physical plausibility. For the semantics evaluation with the CLIP score, we observe that SDS often results in lower semantic correctness, even worse than the initial placement.
This indicates the SDS-generated articulation can sometimes produce nonsensical postures, comparing to the neutral rest pose.
Meanwhile, SDS has a lower score in both non-collision and contact scores, indicating the approach leads to severe penetration and floating that even the SDF loss cannot constrain.
Note that initial placement has no inherent physical meaning, as it is simply positioned according to the specified initial setup. To avoid confusion for clarity, we leave these values blank in the table.

\vspace{\subsecmargin}
\subsection{Ablation Study \& Analysis}
\label{sec:ablation}
\vspace{\subsecmargin}




We conduct ablation studies and analysis to evaluate the effectiveness of our proposed pipeline in synthesizing affordance-aware articulation.

\issue{Bone rotation penalty.} We compare the full version of our method with a variant that removes our proposed Bone Rotation Penalty. The bone rotation penalty is designed to prevent unnatural articulations as mentioned in~\cref{sec:single-view}. The Method without this constraint results in synthesizing implausible postures, as visualized in~\cref{fig:ablation} (a).

\issue{Multi-view fine-grained alignment.} We further validate the effectiveness of the multi-view alignment stage for aligning object positioning and pose with accurate depth. Without updating the model through multi-view alignment stage and relying solely on single-view guidance, the model tends to overfit to the given viewpoint. This results in an unnatural look in other viewpoints, as shown in~\cref{fig:ablation} (b).

\issue{Consistent inpainting of partial denoising grid prior.}
\cref{fig:grid_prior} illustrates the effectiveness of our partial denoising grid prior strategy for achieving consistent inpainting. By partially denoising rendered object images across multiple viewpoints, our approach preserves accurate spatial relationships and object postures across views, leading to stable convergence during articulation synthesis. In contrast, direct inpainting across four separate views fails to ensure semantic coherence, potentially confusing the model and resulting in sub-optimal articulation synthesis.

\issue{Convergence analysis.}
In~\cref{fig:exp-sds-convergence}, we show that our method has a very stable convergence in the multi-view alignment stage.
In comparison, SDS maintains a high stochasticity in bone placement without clear optimization directions.
Considering we have adopted HiFA scheduling, meaning a low noise rate by the end of optimization, the stochasticity remains high and nondeterministic. 
Despite such a stochasticity can be controlled with learning rate decay, but the randomness (\ie moving the bones back-and-forth) in optimization along with learning decay significantly constrains the degree of changes in articulation.


\issue{Out-of-domain rare objects.} 
In \cref{fig:out-of-domain}, we examine the limit of our framework with objects having uncommon appearance and geometry. 
\modelName can still synthesize novel postures with these abnormal examples.
\vspace{\secmargin}
\section{Conclusion}
\label{sec:conclusion}
\vspace{\secmargin}

We propose and tackle a practically valuable task of synthesizing affordance-aware articulation for rigged objects collected from the Internet.
Different from pure object or environment synthesis, our task involves extracting the information of potential object behaviors within an environment.
The task is especially challenging with open-domain objects, the lack of training data, and the naive approaches are all limited by practical challenges.
Our \modelName features solving task with a fast and steady optimization.

\issue{Future work.}
As the recent advancements in video synthesis and increasing number of open-source video diffusion models~\cite{blattmann2023stable,yang2024cogvideox} available, an important next step in our research is distilling the motion in addition to the affordance information.
However, it is more challenging than only synthesizing the articulation, with more information to synthesis and less tools available with video diffusion models.
%
%

\issue{Limitations.}
Despite presented some exciting preliminary results, we still find a few fundamental limitations with \modelName.
We observe that the inpainting diffusion models have a worse text-image alignment, compared to the unconditional diffusion models.
It is still unclear if such a performance gap is purely due to the insufficient training, problems in training paradigm, or a fundamentally unsolvable issue of conditional models.
%
This questions the long-term validity of relying on the inpainting models, and motivates shifting the framework design to the unconditional models.
%

%
%

{
    \small
    \bibliographystyle{ieeenat_fullname}
    \bibliography{main}
}


\clearpage
\maketitlesupplementary
\appendix

\lstset{
    basicstyle=\ttfamily,
    frame=single,
    columns=fullflexible,
    breaklines=true,
    breakautoindent=false,
    breakindent=0ex
}

In the supplementary materials, we provide more implementation details to ensure reproducibility and offer a comprehensive understanding of our proposed framework. We also report additional results for each stage in the framework, such as the matching results of our proposed bone correspondence, and the intermediate steps during multi-view alignment stage. These results verify the contributions of each component and provide deep understanding to the step-by-step refinement process.

\section{Implementation Details}
\label{sec:implementaiton}

We elaborate on our inpainting process in~\cref{sec:inpainting}, including model selection and the corresponding hyper-parameters. We then detail the process of our automatic verification mechanism using Vision-Language Models (VLMs) in~\cref{sec:verification}. We also provide the implementation details for the multi-view alignment stage in~\cref{sec:detail_mv_alignment}. A comprehensive overview of hyper-parameters throughout all experiments and training details are reported in~\cref{sec:parameters}. Lastly, we list the computational resources in~\cref{sec:computational_resources}.

\subsection{Inpainting Details}
\label{sec:inpainting}

The plausibility and diversity of our proposed \modelName{} framework highly rely on the quality of the inpainting process. During the first single-view coarse-grained placement stage, we introduce a state-of-the-art inpainting model, Flux-Controlnet-Inpainting (Flux-CN-Inp.)~\cite{flux_contronlnet_inpainting} to synthesize realistic interaction between objects and scenes. This model is based on the state-of-the-art open-sourced text-to-image generative models, Flux.1-dev~\cite{flux}, and further fine-tuned on a subset of Laion-2B dataset~\cite{schuhmann2022laion} (12M data as reported by the original repository) and internal source images, resulting in realistic quality of interaction between objects and scenes. 

\begin{figure}[t]
    \centering
    \renewcommand{\tabcolsep}{5pt}
    \renewcommand{\arraystretch}{0.5} 
    \begin{tabular}{cc}
        \includegraphics[width=0.45\linewidth]{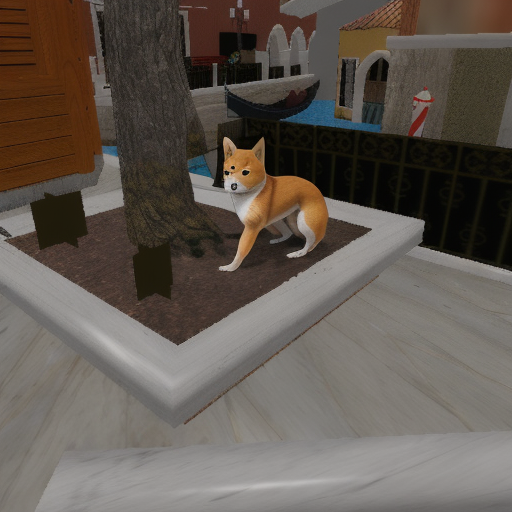} &
        \includegraphics[width=0.45\linewidth]{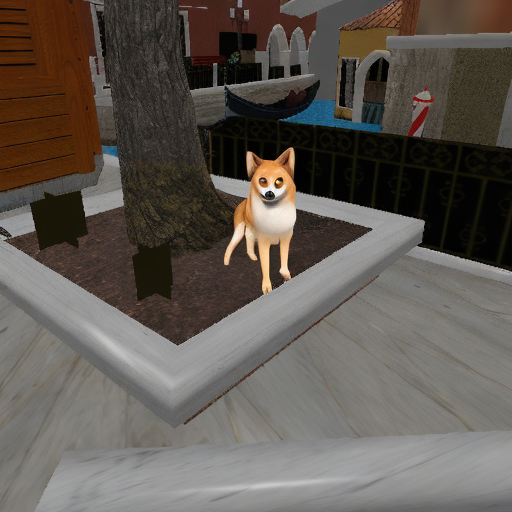} \\
        \vspace{0.5em}
        {\fontsize{8}{4}\selectfont DS-8} &
        {\fontsize{8}{4}\selectfont SD-2} \\

          \includegraphics[width=0.45\linewidth]{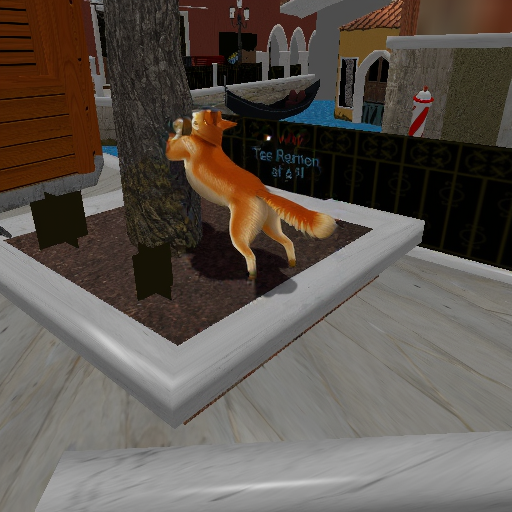} &
        \includegraphics[width=0.45\linewidth]{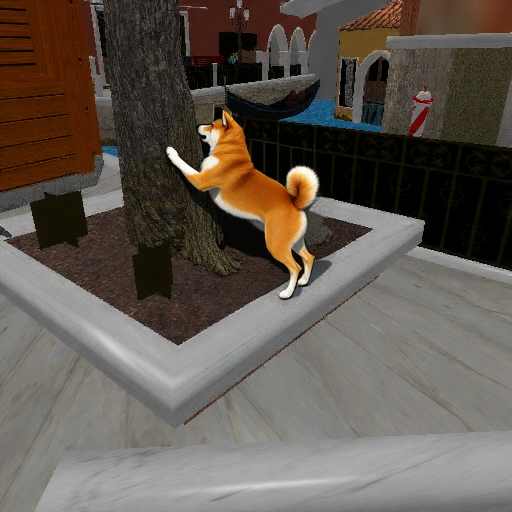}\\

        {\fontsize{8}{4}\selectfont Flux-Inp.} &
        {\fontsize{8}{4}\selectfont Flux-CN-Inp.} \\
    \end{tabular}
    \caption{
        \textbf{Comparison among different inpainting models.} We demonstrate the quality of the inpainted images using different inpainting models. Text prompt: \textit{A shiba inu climbing a tree, with its front paws gripping the rough bark while its hind legs remain planted firmly on the ground for support.} Among four different models, only Flux-Controlnet-Inpainting (Flux-CN-Inp.) produce realistic results, being aware of the interaction between foreground objects and background scene.
    }
    \label{fig:inpainting}

\end{figure}

\cref{fig:inpainting} compares the generated quality across various inpainting models. Besides Flux-CN-Inp., DreamShaper-8-Inpainting (DS-8) is fine-tuned on Stable-Diffusion v1-5 inpainting model~\cite{rombach2022high}. Stable-Diffusion-2-Inpainting (SD-2) is fine-tuned on Stable-Diffusion 2~\cite{rombach2022high}. Flux-Inpainting (Flux-Inp) is an inpainting pipeline using Flux.1-schnell~\cite{flux} with blended latent diffusion~\cite{avrahami2023blended} without additional fine-tuning. The results indicate that both DS-8 and SD-2 fail to align with the provided text prompts, often producing semantically incoherent results. On the other hand, while Flux-Inp. produce an object with the specified action, the details (\eg the head and the front legs) of the generated object are not realistic. Among the models evaluated, Flux-CN-Inp. is the only model producing high-quality and satisfactory results that align with the text prompts, demonstrating realistic object-scene interactions.

\subsection{Verification Process}
\label{sec:verification}

Although we have leveraged the state-of-the-art inpainting model to synthesize high-fidelity 2D object-scene interaction images, the quality is not always satisfactory, and need to be further verified. For real-world applications, we assume that this verification process could be performed by users. However, to automate the entire pipeline, we also explore the potential of advanced Vision-Language Models (VLMs) to assess the plausibility of inpainted images and evaluate the alignment between the generated images and their corresponding text prompts.

We leverage GPT-4o~\cite{openai2024chatgpt} to help verify the quality of the inpainted images generated during the first stage. We prompt the model using the following instruction to evaluate the alignment between images and text prompts.

\begin{lstlisting}[breaklines=true]
Given an image, evaluate whether the posture of the foreground object and its interaction with the background align with the provided prompt: "<PROMPT>". If the image aligns correctly with the prompt, return: ```json{"is_valid": true}```, otherwise return ```json{"is_valid": false}```

\end{lstlisting}

\begin{figure}[t]
    \centering
    \renewcommand{\tabcolsep}{5pt}
    \renewcommand{\arraystretch}{0.5} 
    \begin{tabular}{cc}
        \includegraphics[width=0.45\linewidth]{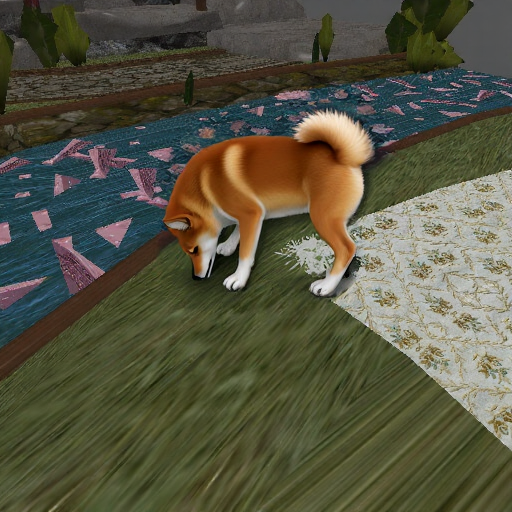} &
        \includegraphics[width=0.45\linewidth]{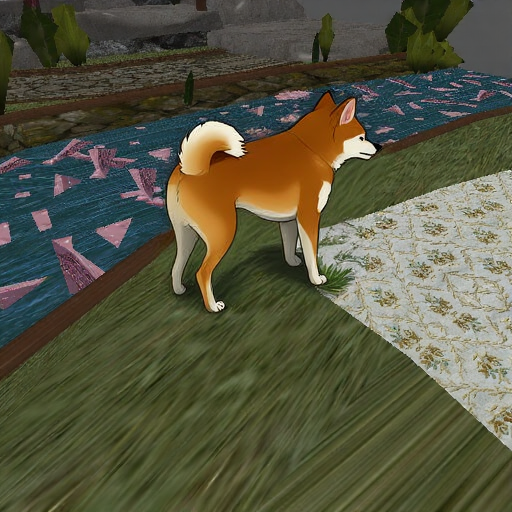} \\
        {\fontsize{8}{4}\selectfont is\_valid: true} &
        {\fontsize{8}{4}\selectfont is\_valid: false}\\
    \end{tabular}
    \caption{
        \textbf{Automatic VLMs verification process for the inpainted images.} 
        We leverage advanced VLMs to automatically verify the quality of inpainted images and evaluate their alignment with the provided text prompts. In this example, the text prompt is: \textit{ A Shiba Inu standing in a grassy field, lowering its head to graze on fresh green grass}. \textbf{Left.} An example of an image that successfully passes the verification. \textbf{Right.} An example of an image that fails the verification process. 
    }
    \label{fig:verification}

\end{figure}

We replace $<$PROMPT$>$ with the specific prompt used to generate the inpainted image. The output is a JSON-formatted boolean value indicating whether the inpainted image passes GPT-4o's verification process. \cref{fig:verification} provides examples of a successfully verified image and an image that fails the verification. In the failed example, the object is merely inpainted into the scene without performing the action specified in the prompt. If an image fails the verification, we simply select a new seed, generate a new inpainted image, and repeat the verification process until a satisfactory result is achieved.

\subsection{Details of Multi-View Fine-Grained Alignment}
\label{sec:detail_mv_alignment}
\begin{figure}[t]
    \centering
    \renewcommand{\tabcolsep}{5pt}
    \renewcommand{\arraystretch}{0.5} 
    \begin{tabular}{cc}
        \includegraphics[width=0.45\linewidth]{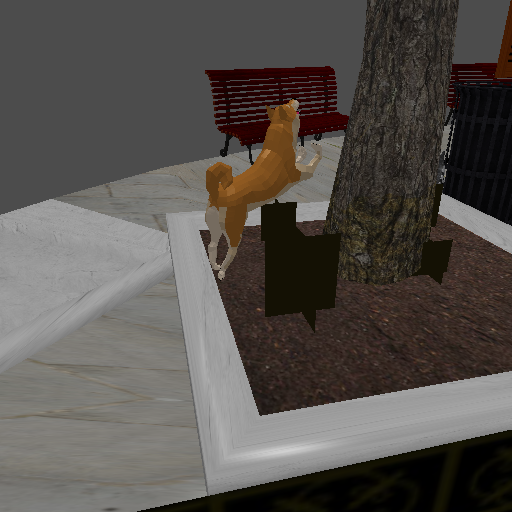} &
        \includegraphics[width=0.45\linewidth]{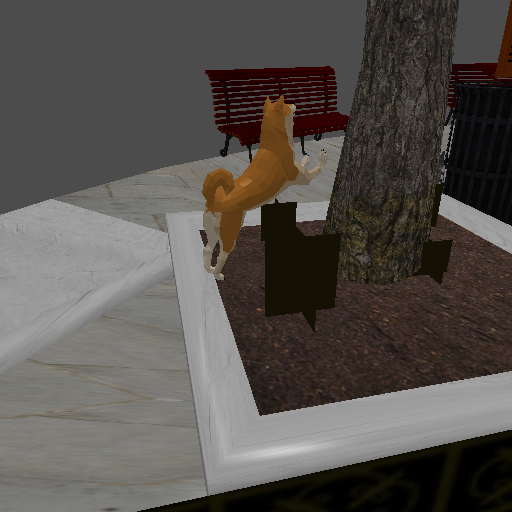} \\
        \vspace{0.5em}
        {\fontsize{8}{4}\selectfont Before MV alignment} &
        {\fontsize{8}{4}\selectfont 1-st round MV alignment} \\

          \includegraphics[width=0.45\linewidth]{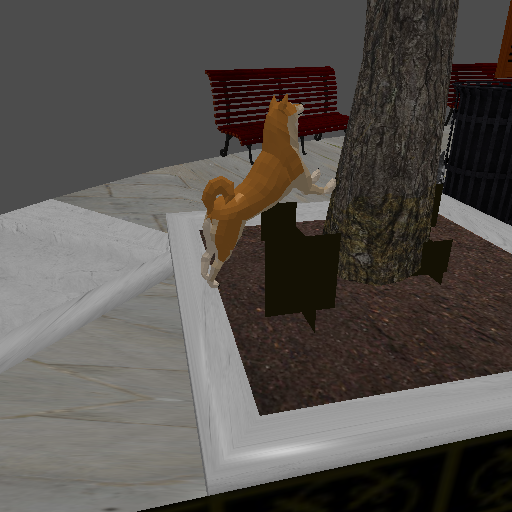} &
        \includegraphics[width=0.45\linewidth]{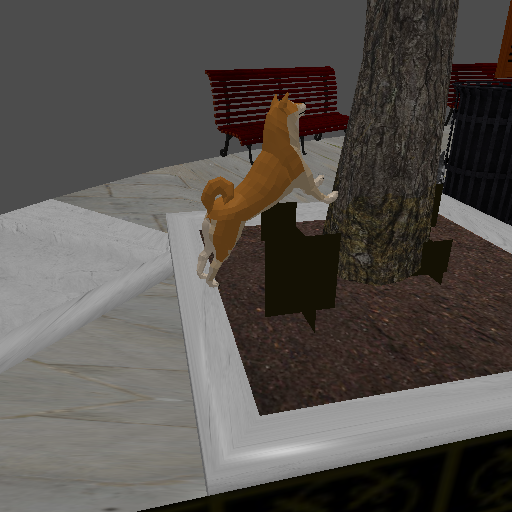}\\

        {\fontsize{8}{4}\selectfont 2-nd round MV alignment} &
        {\fontsize{8}{4}\selectfont 3-rd round MV alignment} \\
    \end{tabular}
    \caption{
        \textbf{Effectiveness of Iterative Refinement of Multi-view fine-grained alignment.} We demonstrate the effectiveness of multi-view fine-grained alignment at each stage. The depth ambiguity issue is iteratively improved after each iteration of multi-view alignment.
    }
    \label{fig:mv_alignment}

\end{figure}

During the multi-view fine-grained alignment stage, we choose Flux-Inpainting as our backbone inpainting models, as the implementation of blended latent diffusion naturally supports partial denoising. Across the whole experiments, we iterate the multi-view alignment process three times, with descending partial denoising ratios $\tau$ from $0.8$ to $0.6$. Such the descending ratios are helpful for the convergence of the pipeline, reducing the possibility that the inpainting pipeline generates completely different reference images every iteration. \cref{fig:mv_alignment} shows the effectiveness of our proposed multi-view alignment at each iteration. As the number of iterations increases, the synthesized articulation is converged, and the orientation is getting more natural.

\begin{figure}[t]
    \centering
    \renewcommand{\tabcolsep}{1pt}
    \renewcommand{\arraystretch}{0.5} 
    \begin{tabular}{lcccc}
    \parbox[c]{.8em}{\rotatebox[origin=c]{90}{\small Rendered\hspace{-5em}}} &
        \includegraphics[width=0.23\linewidth]{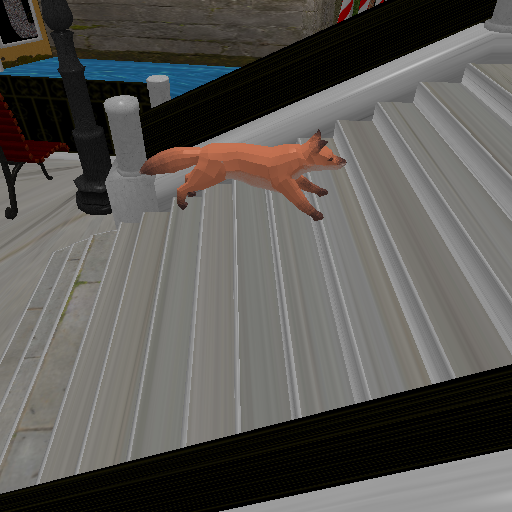} &
         \includegraphics[width=0.23\linewidth]{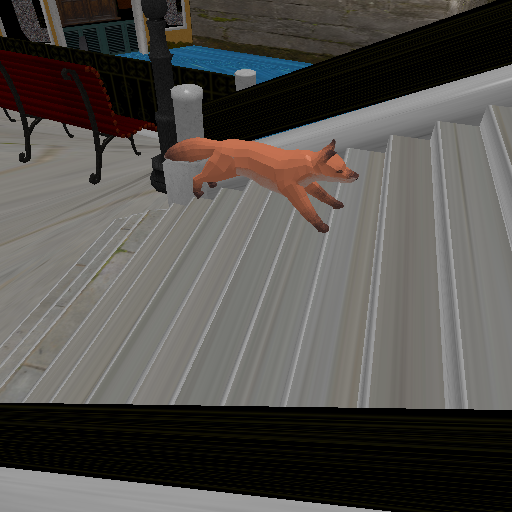} &
          \includegraphics[width=0.23\linewidth]{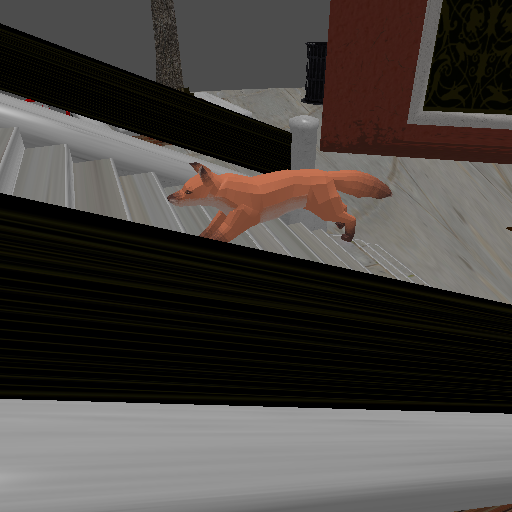} &
        \includegraphics[width=0.23\linewidth]{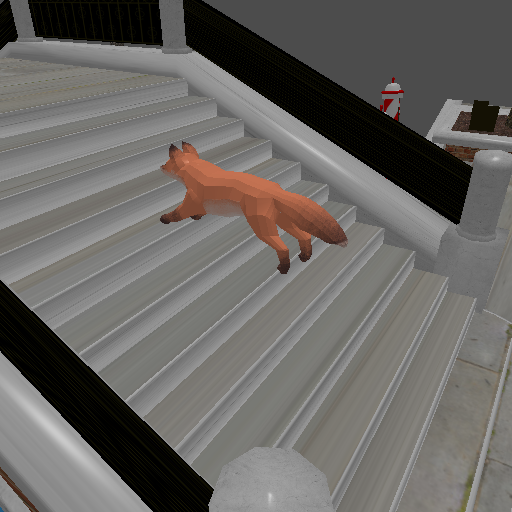} \\
        \parbox[c]{.8em}{\rotatebox[origin=c]{90}{\small Inpainted\hspace{-5em}}} &
        \includegraphics[width=0.23\linewidth]{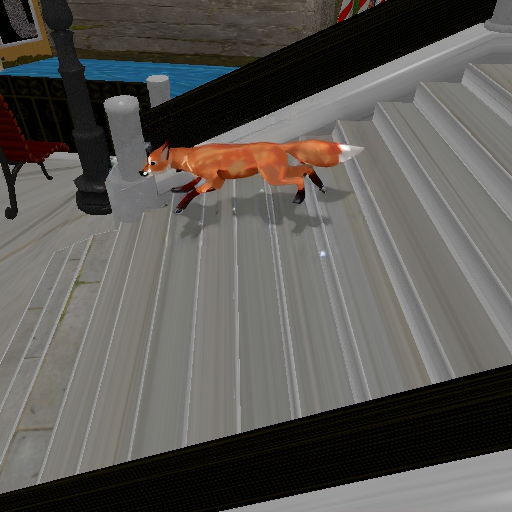} &
         \includegraphics[width=0.23\linewidth]{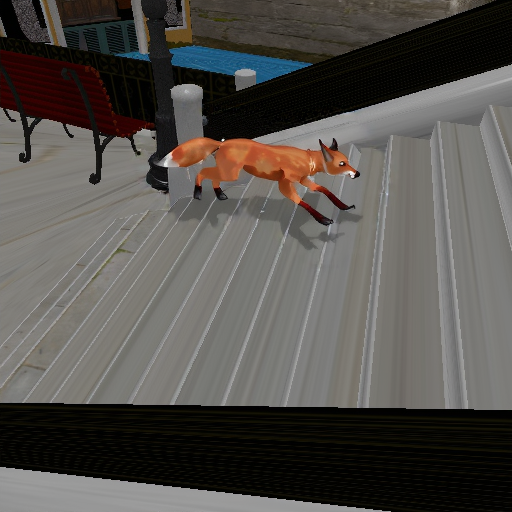} &
          \includegraphics[width=0.23\linewidth]{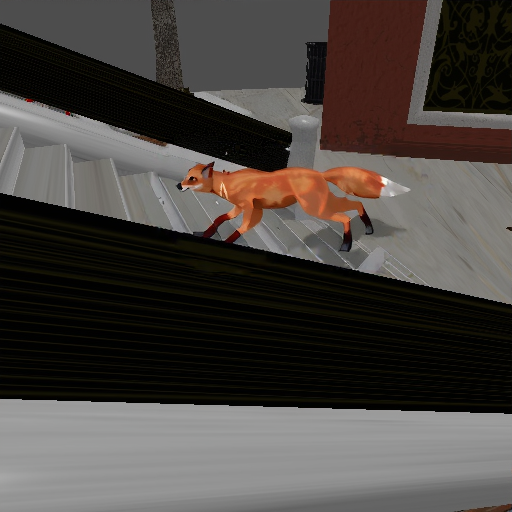} &
        \includegraphics[width=0.23\linewidth]{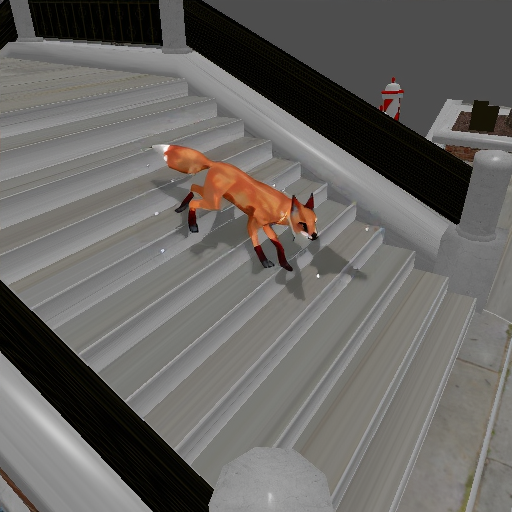} \\
        &
        {\fontsize{8}{4}\selectfont view 0} &
        {\fontsize{8}{4}\selectfont view 1} &
        {\fontsize{8}{4}\selectfont view 2} &
        {\fontsize{8}{4}\selectfont view 3} \\
    \end{tabular}
    \caption{
        \textbf{Visualization of the reverse head-to-tail postures.} 
        When the partial denoising rate is higher (\eg $0.8$, in this example), the inpainted model sometimes generates objects with reverse head-to-tail postures. In this example, the inpainted images on view $0$ and view $3$ are in reverse postures, resulting in abnormally high correspondence loss. Thus, we apply a simple loss threshold $\epsilon_t$ to ignore the huge loss derived from these viewpoints.
    }
    \label{fig:loss_threshold}

\end{figure}

\issue{Multi-view loss threshold.}
Although the inpainted image quality from different viewpoints are mostly consistent due to our proposed partial denoising grid-prior strategy. Empirically, we find the inpainting model sometimes produces objects with reverse head-to-tail postures, as demonstrated in~\cref{fig:loss_threshold} In this case, the bone correspondence loss $\calL_{\text{BC}}^m$ will be abnormally larger than usual. Thus, we apply a simple loss threshold $\epsilon_t$ to exclude anomalous loss values, as mentioned in~\cref{sec:multi-view}. We set $\epsilon_t$ as $1000$ across the whole experiments. When there are no images provided valid loss range, we simply ignore this round, and re-inpaint a set of new multi-view images. 

\subsection{Other Implementation Details}
\label{sec:parameters}

\issue{Decomposition of the SDF Loss.}
The SDF Loss $\calL_{\text{SDF}}$ was originally proposed by~\cite{li2024genzi}. It can be decomposed into two components: the penetration loss $\calL_{\text{pen}}$ and the no-contact loss $\calL_{\text{no-cont}}$, defined as follows:

\begin{align}
&\calL_{\text{pen}} & &= \sum_{v \in V} ||\min(\Psi(v), 0)||_1, \\
&\calL_{\text{no-cont}} & &= \min_{v \in V} \Psi(v),
\end{align}
and the original SDF loss $\calL_{\text{SDF}}$ can be expressed as:

\begin{equation}
\calL_{\text{SDF}} = \lambda_{\text{pen}} \cdot \calL_{\text{pen}} + \lambda_{\text{no-cont}} \cdot \calL_{\text{no-cont}},
\end{equation}
where $\lambda$ are hyper-parameters to control the balance among different loss functions.

As discussed in~\cref{sec:single-view}, when all vertices in $V$ have positive SDF values, we calculate $\calL_{\text{no-cont}}$ using the above definition and set $\calL_{\text{pen}} = 0$. Conversely, if any vertex has a negative SDF value, we calculate $\calL_{\text{pen}}$ and set $\calL_{\text{no-cont}} = 0$.

\issue{Hyper-parameters for rotation penalty loss.}
The base $\alpha$ in~\cref{eq:rotation_penalization} controls the growth rate of the penalty applied to the rotation of child nodes. As $\alpha$ increases, rotations of child nodes become increasingly restricted. However, higher $\alpha$ can also excessively limit the diversity of synthesized articulations. To balance naturalness and diversity in synthesized postures, we set $\alpha = 1.2$ for all experiments.

\issue{Hype-parameters for loss functions.}
The weights $\lambda$ for each loss function are critical for reflecting their relative importance. Empirically, we find the range of different loss functions vary significantly. For example, the SDF loss is typically 1000x smaller than the bone correspondence loss. To normalize the impact of each loss, we set $\lambda_{\text{BC}} = \lambda_{\text{MVBC}} = 1$, $\lambda_{\text{RP}} = 100$, and $\lambda_{\text{pen}} = \lambda_{\text{no-cont}} = 1000$. These weights ensure that each loss function contributes proportionally to the overall optimization objective. We use the same weights for all experiments.

\issue{Training details.}
We set the learning rate for all learnable parameters (\ie articulation parameters $\mathcal{A}$ and global transformation $\mathcal{T}$) as $10^{-2}$, except for the scale parameter. Empirically, the learning rate for the scale parameter is set to $10^{-5}$ to prevent unnatural scaling of the object. During the single-view coarse-grained placement stage, we update the parameters for a total of 200 epochs to establish an initial alignment. For the multi-view fine-grained alignment stage, We reduce the total number of epochs to 100 since the update in this stage primarily involves subtle adjustments to refine the object postures.
\subsection{Computational Resources}
\label{sec:computational_resources}
All experiments could be done using single RTX A6000 GPU with 48GB memory. 

\section{Intermediate Results}

In this section, we present intermediate results from our experiments, including the matching results of bone correspondence and the intermediate steps of the multi-view alignment stage. These results provide a deeper understanding of the role of each component, and its contribution to the overall performance.

\begin{figure}[t]
    \centering
    \renewcommand{\tabcolsep}{1pt}
    \renewcommand{\arraystretch}{0.5} 
    \begin{tabular}{lcccc}
    \parbox[c]{.8em}{\rotatebox[origin=c]{90}{\small Inpainted (ref.)\hspace{-5em}}} &
        \includegraphics[width=0.23\linewidth]{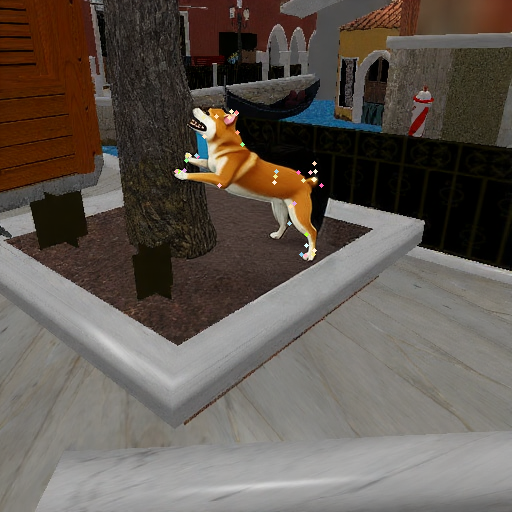} &
         \includegraphics[width=0.23\linewidth]{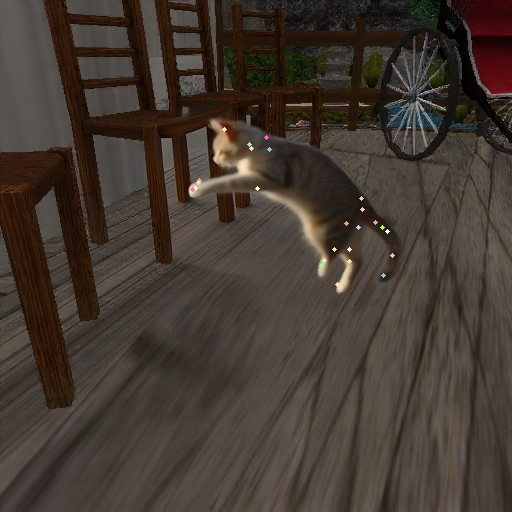} &
          \includegraphics[width=0.23\linewidth]{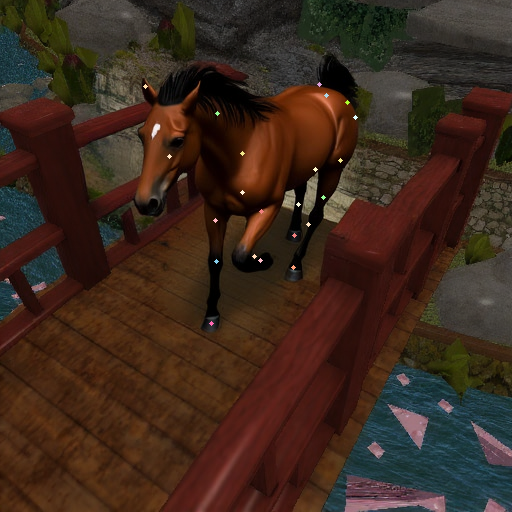} &
        \includegraphics[width=0.23\linewidth]{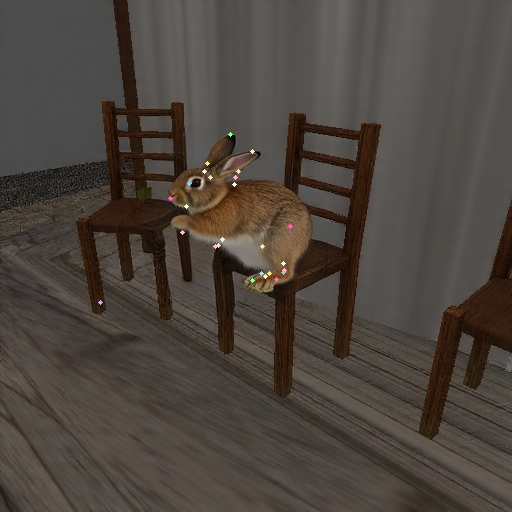} \\
        \parbox[c]{.8em}{\rotatebox[origin=c]{90}{\small Rest pose\hspace{-5em}}} &
        \includegraphics[width=0.23\linewidth]{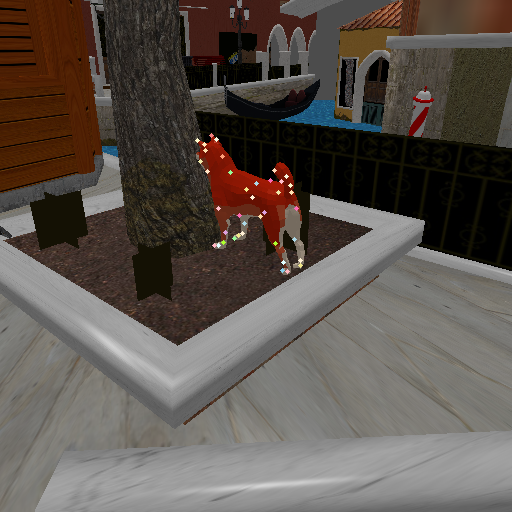} &
         \includegraphics[width=0.23\linewidth]{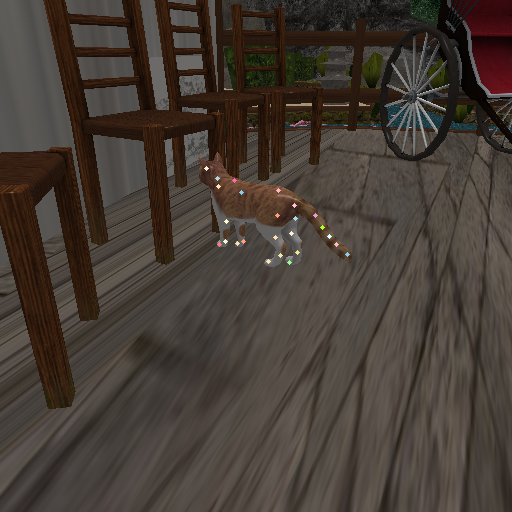} &
          \includegraphics[width=0.23\linewidth]{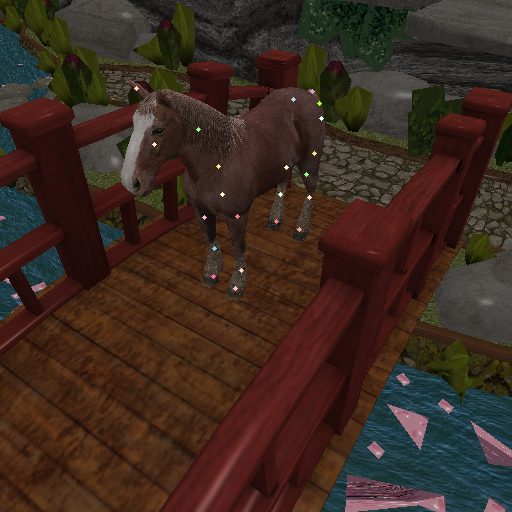} &
        \includegraphics[width=0.23\linewidth]{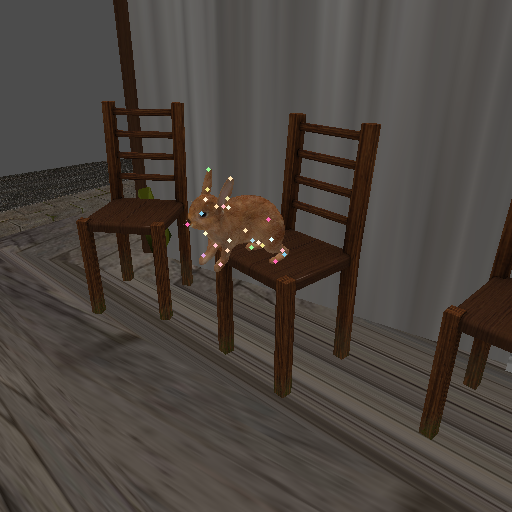} \\

        \parbox[c]{.8em}{\rotatebox[origin=c]{90}{\small Aligned pose\hspace{-5em}}} &
        \includegraphics[width=0.23\linewidth]{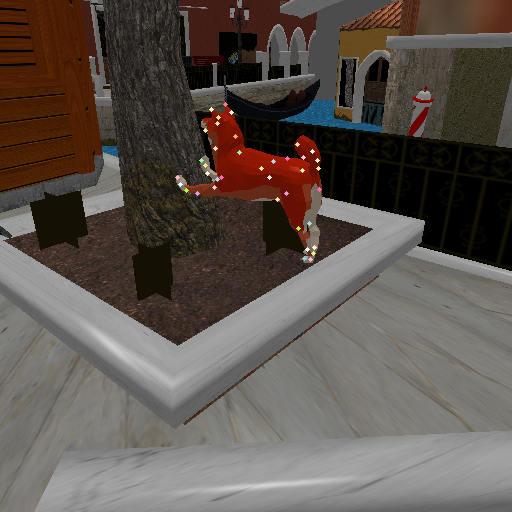} &
         \includegraphics[width=0.23\linewidth]{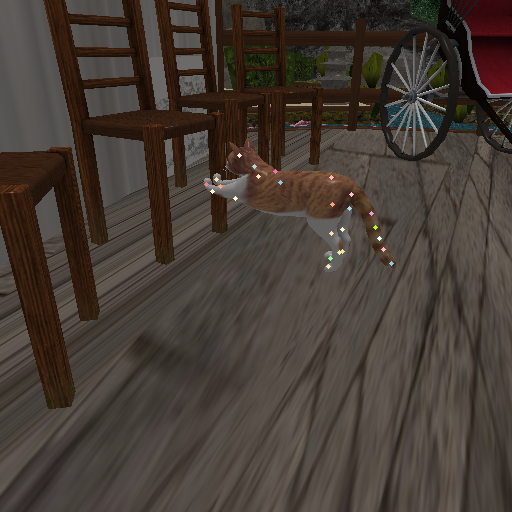} &
          \includegraphics[width=0.23\linewidth]{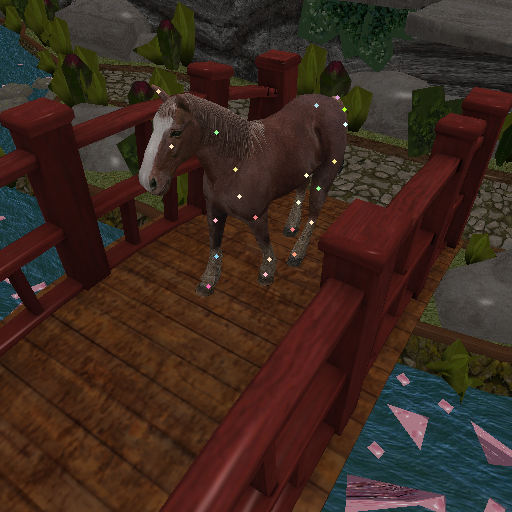} &
        \includegraphics[width=0.23\linewidth]{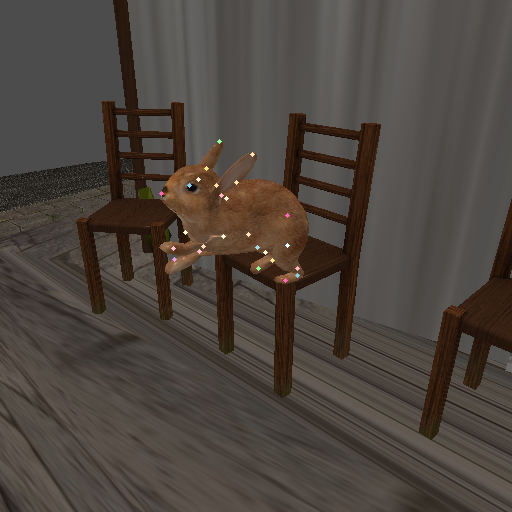} \\
        &
        {\fontsize{8}{4}\selectfont climb tree} &
        {\fontsize{8}{4}\selectfont jump ground} &
        {\fontsize{8}{4}\selectfont walk bridge} &
        {\fontsize{8}{4}\selectfont jump chair} \\
    \end{tabular}
    \caption{
        \textbf{Bone correspondence matching results for single-view coarsed-grained placement.} The inpainted image serves as a reference, providing the desired posture for alignment. Each color indicates bone correspondence points across the query image and the reference image. Our proposed bone correspondence effectively matches the bones in 2D space, enabling the synthesis of the desired articulation.
    }
    \label{fig:supp_bc_sv}

\end{figure}

\issue{Bone correspondence for single-view placement.}
Our proposed bone correspondence loss aims to align the visible bone positions between the query image (rendered image) and the target image (inpainted image) from a single viewpoint. The effectiveness of this method heavily depends on the accuracy of the correspondence points. Accurate matching ensures that the synthesized articulation aligns closely with the desired posture in the target image. \cref{fig:supp_bc_sv} illustrates the intermediate results of the bone correspondence matching process, providing reliable guidance for the subsequent articulation synthesis.

\begin{figure}[t]
    \centering
    \renewcommand{\tabcolsep}{1pt}
    \renewcommand{\arraystretch}{0.5} 
    \begin{tabular}{lcccc}
    \parbox[c]{.8em}{\rotatebox[origin=c]{90}{\small Init. pose\hspace{-5em}}} &
        \includegraphics[width=0.23\linewidth]{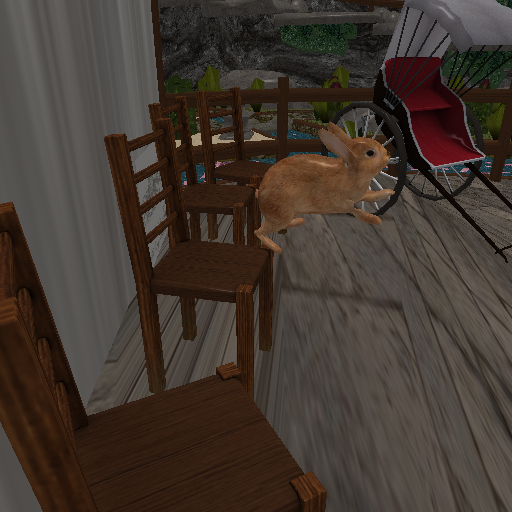} &
         \includegraphics[width=0.23\linewidth]{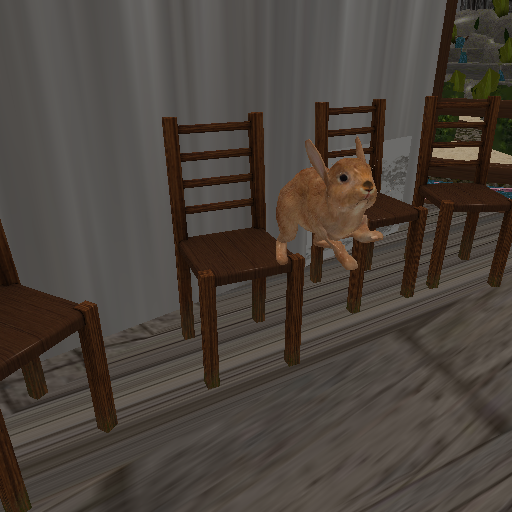} &
          \includegraphics[width=0.23\linewidth]{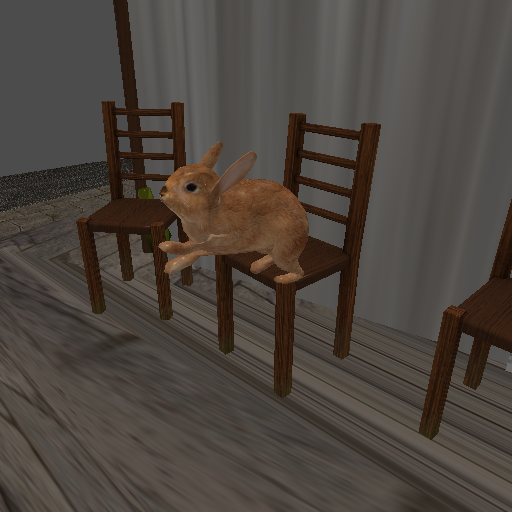} &
        \includegraphics[width=0.23\linewidth]{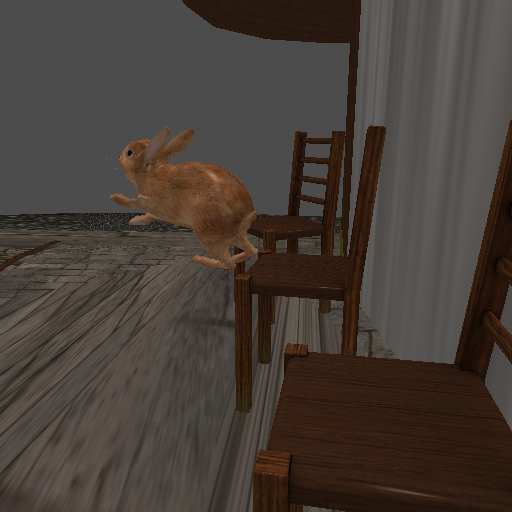} \\
        \parbox[c]{.8em}{\rotatebox[origin=c]{90}{\small Round 1\hspace{-5em}}} &
        \includegraphics[width=0.23\linewidth]{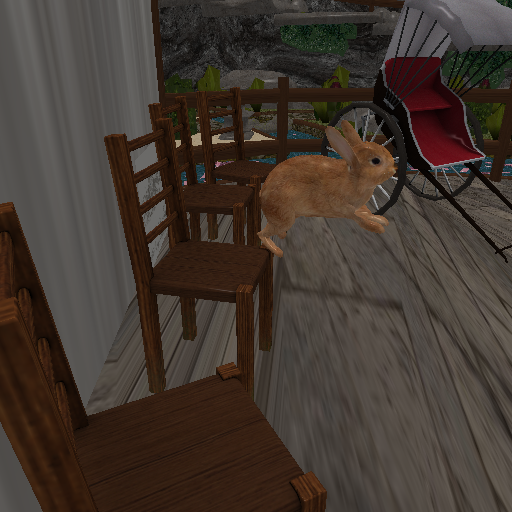} &
         \includegraphics[width=0.23\linewidth]{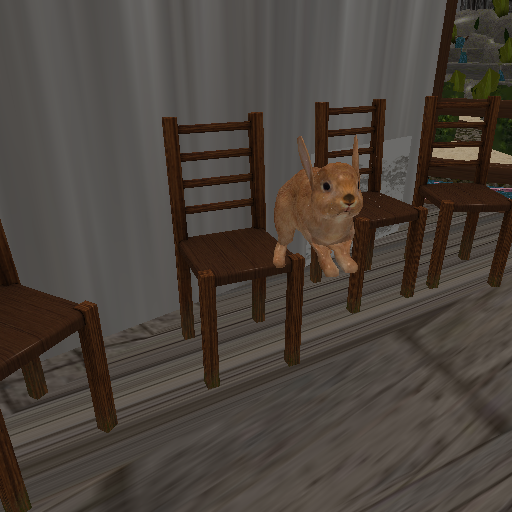} &
          \includegraphics[width=0.23\linewidth]{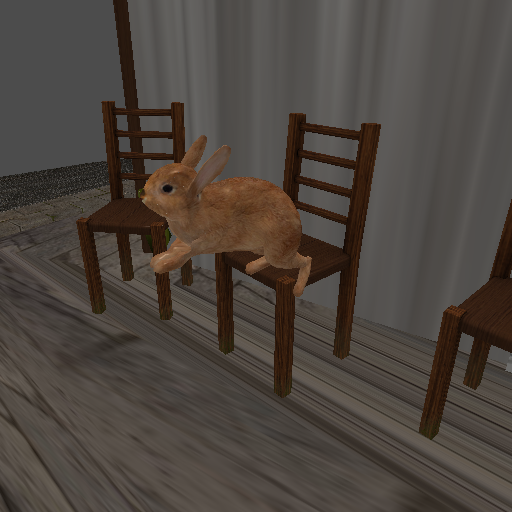} &
        \includegraphics[width=0.23\linewidth]{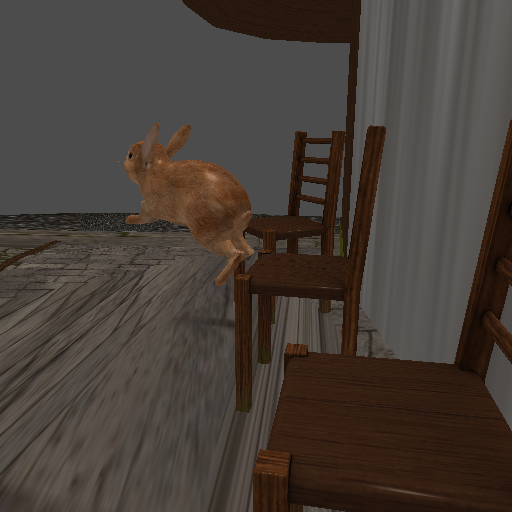} \\

        \parbox[c]{.8em}{\rotatebox[origin=c]{90}{\small Round 2\hspace{-5em}}} &
        \includegraphics[width=0.23\linewidth]{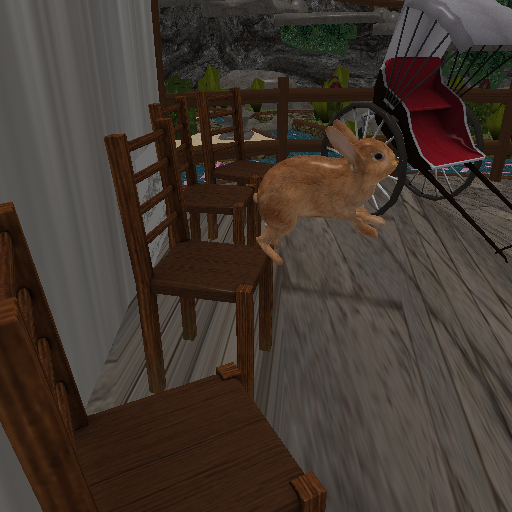} &
         \includegraphics[width=0.23\linewidth]{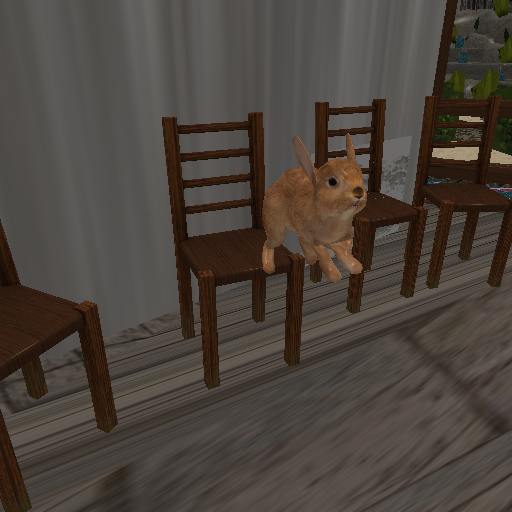} &
          \includegraphics[width=0.23\linewidth]{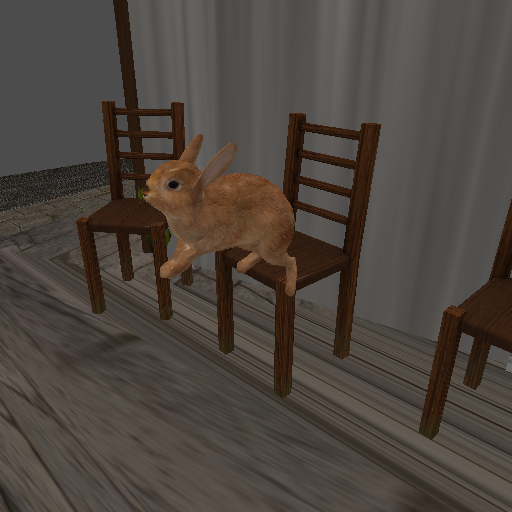} &
        \includegraphics[width=0.23\linewidth]{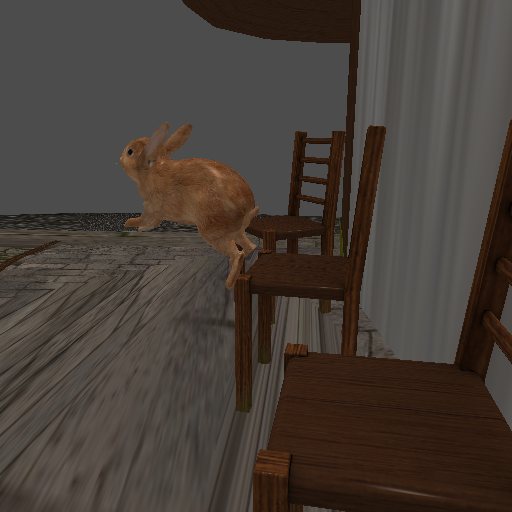} \\
        \parbox[c]{.8em}{\rotatebox[origin=c]{90}{\small Round 3\hspace{-5em}}} &
        \includegraphics[width=0.23\linewidth]{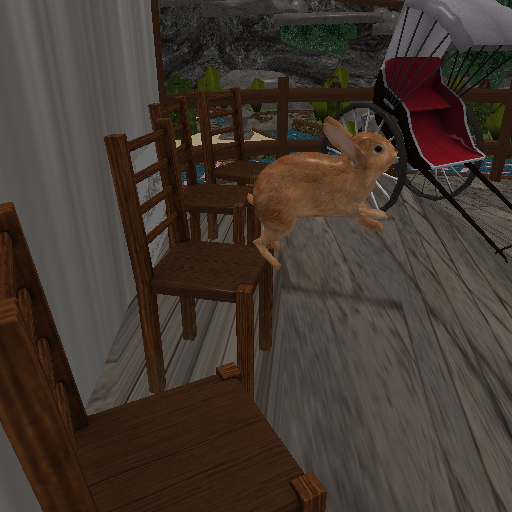} &
         \includegraphics[width=0.23\linewidth]{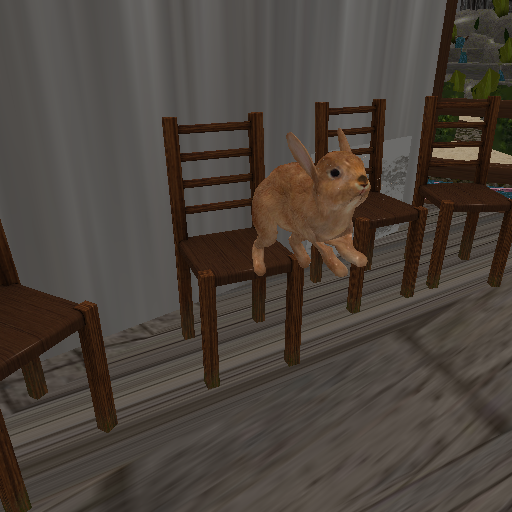} &
          \includegraphics[width=0.23\linewidth]{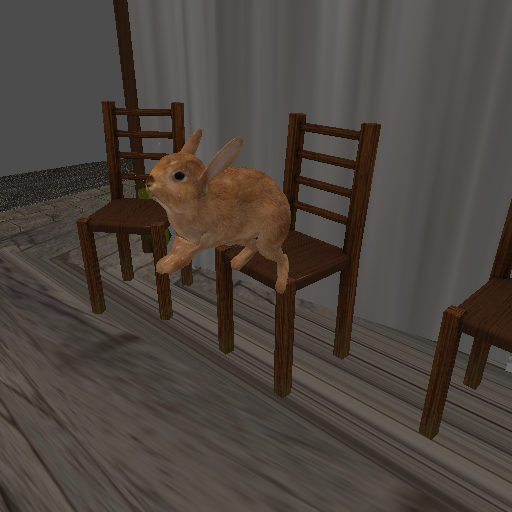} &
        \includegraphics[width=0.23\linewidth]{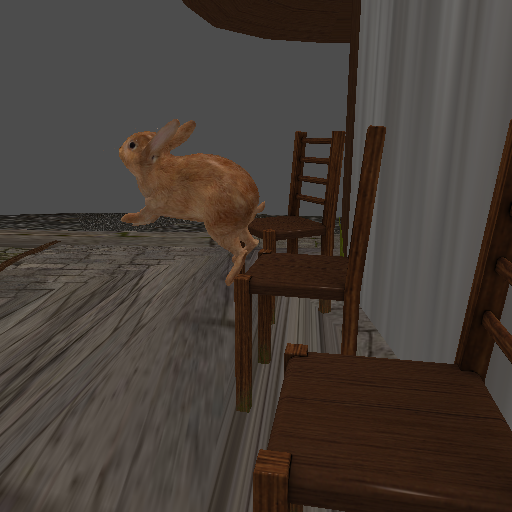} \\
        &
        {\fontsize{8}{4}\selectfont View 0} &
        {\fontsize{8}{4}\selectfont View 1} &
        {\fontsize{8}{4}\selectfont View 2} &
        {\fontsize{8}{4}\selectfont View 3} \\
    \end{tabular}
    \caption{
        \textbf{Intermediate steps for multi-view fine-grained alignment stage.} Text prompt: \textit{A brown rabbit in mid-leap as it jumps down from a wooden chair.} In this example, the left hind leg of the rabbit are wrongly placed after the single-view placement stage (initial pose), appearing in an unnatural position. During the multi-view alignment stage, the posture is iteratively refined, with the left hind leg gradually adjusting to align more realistically with the action described in the prompt.
    }
    \label{fig:supp_mv_intermediate}

\end{figure}

\issue{Intermediate steps during MV alignment.}
The multi-view fine-grained alignment stage primarily corrects biases introduced during the initial single-view placement stage. Additionally, it also making subtle adjustments to the object's posture. \cref{fig:supp_mv_intermediate} presents an example of the intermediate results at each round of the multi-view alignment stage. In this example, the left hind leg of the rabbit is incorrectly positioned in the initial pose (\ie the result from the single-view placement stage). However, it is gradually adjusted during the multi-view alignment process, finally achieving a more realistic posture when viewed from multiple angles. This verifies the effectiveness of the multi-view iterative alignment stage in refining object postures, improving the overall plausibility of the synthesized articulation across different viewpoints.

\section{License}

We thank all the following artists for creating the 3D objects used in our work and generously shared them for free on Sketchfab.com: ``\href{https://skfb.ly/6TKqH}{3d modelling my cat: Fripouille}'' by guillaume bolis, ``\href{https://skfb.ly/6nVN9}{Cow NPC}'' by Owlish Media, ``\href{https://skfb.ly/p8wzz}{Horse Rigged(Game Ready)
}'' by abhayexe, ``\href{https://skfb.ly/o6Joz}{Low poly fox running animation}'' by dragonsnap, ``\href{https://skfb.ly/69ps6}{Shiba Inu Doggy}'' by aaadragon, ``\href{https://skfb.ly/6V6FS}{Rabbit Rigged}'' by FourthGreen, ``\href{https://skfb.ly/oFCQX}{Spongebob. Rigged}'' by Eyeball, ``\href{https://skfb.ly/oFCRw}{Patrick. Rigged}'' by Eyeball, ``\href{https://skfb.ly/6TptH}{Venice city scene 1DAE08 Aaron Ongena}'' by AaronOngena, ``\href{https://skfb.ly/6ToLn}{1DAE10 Quintyn Glenn City Scene Kyoto}'' by Glenn.Quintyn, ``\href{https://skfb.ly/6QYJI}{Low Poly Farm V2} by EdwiixGG. All 3D objects are licensed under CC Attribution.

\end{document}